%% file: Concordia_Univ_Seminar_paper.tex
\documentclass[11pt,a4paper]{article}
\usepackage[utf8]{inputenc}
\usepackage{amsmath}
\usepackage{amsfonts}
\usepackage{amssymb}

\date{}

\usepackage{graphicx,xspace,color}

\graphicspath{{/home/sfgk/2024/Figs/pinn_figs/},{/home/sfgk/2024/Figs/},{/home/sfgk/2024/Figs/Png/},{/home/sfgk/2024/Figs/Pdf/},{/home/sfgk/2024/Figs/img/},{/home/sfgk/2024/Figs/Figs0/},{/home/sfgk/2024/Figs/Figs1/},{/home/sfgk/2024/Figs/Figs2/},{/home/sfgk/2024/Figs/Figs3/},,{/home/sfgk/2024/Figs/Figs4/},{/home/sfgk/2023/Images/}{/home/sfgk/2024/Figs1/},{/home/sfgk/2022/Figs1/},{/home/sfgk/2022/Figs2/},{/home/sfgk/2022/Figs3/},{/home/sfgk/2022/Pdf/},{/home/sfgk/2023/Figs/},{/home/sfgk/2022/Pdf/}}

\usepackage{tikz}
\usetikzlibrary{decorations.pathreplacing}

\def\REM#1{}

\input{\inputdir alphabet1}

\input{\inputdir alphabet2}

\input{\inputdir macros_gpi}

\def\rfbnn{\rfb_{\tiny NN}}
\def\bvi{\blue{v_{i}}}

\begin{document}

\title{Bayesian Physics Informed Neural Networks for Linear Inverse problems}

\author{A. Mohammad-Djafari\\ ~\\ 
former Research Director at CNRS, CentraleSup\'elec, Gif-sur-Yvette, France. 
}
\maketitle

\begin{abstract}
Inverse problems arise almost everywhere in science and engineering where we need to infer on a quantity from indirect observation. The cases of medical, biomedical, and industrial imaging systems are the typical examples.
\\
A very high overview of classification of the inverse problems method can be: i) Analytical, ii) Regularization, and iii) Bayesian inference methods. 
Even if there are straight links between them, we can say that the Bayesian inference based methods are the most powerful, as they give the possibility of accounting for prior knowledge and can account for errors and uncertainties in general. One of the main limitations stay in computational costs in particular for high dimensional imaging systems. Neural Networks (NN), and in particular Deep NNs (DNN), have been considered as a way to push farther this limit. 
\\
Physics Informed Neural Networks (PINN) concept integrates physical laws with deep learning techniques to enhance the speed, accuracy and efficiency of the above mentioned problems. 
\\ 
In this work, a new Bayesian framework for the concept of PINN (BPINN)  is presented and discussed which includes the deterministic one if we use the Maximum A Posteriori (MAP) estimation framework. We consider two cases of supervised and unsupervised for training step, obtain the expressions of the posterior probability of the unknown variables, and deduce the posterior laws of the NN's parameters. We also discuss about the challenges of implementation of these methods in real applications. 
 
\end{abstract}

\section{Introduction}
Inverse problems arise in many scientific and engineering domains, such as medical and biological imaging \cite{tomo-MI}, geophysical imaging \cite{Seismic-tomo}, industrial imaging \cite{Inspection-IP}, and many others \cite{e23121673}.  
These problems aim to infer an inaccessible quantity (input parameters of a system), such as physical fields, source locations, initial and boundary conditions, from the corresponding indirect measurement data. Inverse problems are, in general, ill-posed, as defined by Hadamard: 
A mathematical problem is saied to be well-posed, if it has a solution (Existance), if the solution is unique (Uniqueness) and if the solution is stable (stability). If any of these conditions are not satisfied, the problem is qualified as to ill-posed. 

Classical approaches to inverse problems can be classified as three main categories: Analytical, regularization methods and Bayesian inference. Regularization techniques aim to determine point estimates by simultaneously minimizing the misfits between observed and predicted outputs while penalizing undesired parameter features or encourage the desired parameter features through a regularization term \cite{engl1996regularization,SampledTikhonov}. 

Bayesian inference incorporates uncertainties in prior information and observations through the Bayes rule. It constructs the posterior distribution of parameters, thus effectively quantifying the uncertainty of unknown desired variables. However, in many applications, it is computationally expensive to sample from the Bayesian posterior distributions by using statistical simulation techniques, such as the Markov Chain Monte Carlo (MCMC)\cite{MultiscaleMRM, DataDriven,InvStekloff,IP-NU}, Hamiltonian Monte Carlo \cite{HMC-tomo,DD-HMC}, and Sequential Monte Carlo \cite{Multicale-SMC,SMC}. 

To address this issue, many researchers utilized various variational inference algorithms to reduce the computational cost, including mean-field variational inference \cite{ApproximateMI,BVI-Flow}, automatic differential variational inference\cite{VI-tomo,ADVI-Geo}, and Stein variational gradient descent\cite{VI-tomo,SVGD-inv}. But the variational methods may face challenges in high-dimensional settings due to the difficulty of accurately approximating posterior distributions with the tractable surrogate distribution. 

Although, even if all aforementioned methods have shown to be relatively efficient, they still require numerous evaluations of the forward model and the complicated parametric derivatives, which introduce higher computational cost in high-dimensional real applications. Consequently, deep learning neural networks  based methods have became an efficient alternative, where they are capable of providing almost real-time inversions for certain classes of inverse problems with new measurement data of similar type, under the conditions that they have been  trained properly with a great number of good quality training data.  

Deep learning methods have emerged as a promising approach for fast computation of the solutions of inverse problems, such as medical imaging \cite{CNN-IP-image,Deep-MIA} and electromagnetic inversion \cite{Deep-EI,Deep-EI-1d}. Nevertheless, these methods heavily rely on labeled data from solutions of the forward problem, rendering them inappropriate for inverse problems where such information is not present.

Physics-informed neural network (PINN) methods have been developed to regularize and generalize Deep learning methods by accounting for physical properties of the training data. These methods started to be used for inverse problems described by the ordinary or partial differential equations (ODE or PDE) \cite{PINN}. 
Since the starting point in 2017, \cite{Raisi}, extensive researches have demonstrated the efficiency of the concept, mainly for ODE and PDE models, both forward and inverse problems \cite{Quantify-PINN,PINN-solid,DeepXDE,B-PINN,Adver-PINN,PINNtomo,PINN-wavefield,A-PINN,Surrogate-PINN,psf2023009014}. However, their real efficiency for inverse problems have still to be proved. 

However, models based on invertible neural networks (INN) \cite{NICE, RNVP, Glow} have shown potential in Bayesian inference for inverse problems, due to their efficiency and accuracy advantages in both sampling and density estimation of complex target distributions through bijective transformations, as evidenced in the works of \cite{INN, INN-geo, cINN, BayesFlow, NFFs, Inv-DeepONets, L-HYDRA, VI-NFs}. However, still, certain limitations persist. 
For example, INN models in \cite{INN, INN-geo, cINN, BayesFlow} demand substantial quantities of labeled data for pre-training, which is often unavailable in many practical inverse problems. The invertible DeepONet introduced in \cite{Inv-DeepONets} alleviates the need for labeled data, but it entails an additional variational inference procedure to approximate the true posterior distribution.

In this paper, we focus on the Bayesian inference tool, and in a synthetic way, step by step, show the different steps of assigning prior probability distribution, and obtaining the expression of the posterior laws, first for the the desired unknowns, and the for the NN's parameters during the training. We see how we obtain the concept of PINN methods as special cases. We consider two cases of supervised and unsupervised cases in details. 


\section{Proposed Bayesian PINN framework}
We introduce the Bayesian PINN framework by following five steps: 
\bit
\item 
First step is to assume that we have a forward or generative model:
\beq
\bgb=\Hb(\rfb)+\epsilonb
\eeq
where $\rfb$ is the input, $\bgb$ the output, $\Hb$ the forward model, and $\epsilonb$ the errors, both measurement and modeling. For a linear model, we have: $\bgb=\Hb\rfb+\epsilonb$. 

\item 
Second step is to assign priors $p(\rfb)$ and $p(\epsilonb)$. From this second one, and the forward model, we can deduce the likelihood 
$p(\bgb|\rfb)$. Then, we can use the Bayes rule to obtain:  
\beq
p(\rfb|\bgb)=\frac{p(\bgb|\rfb) p(\rfb)}{p(\bgb)}
\eeq
For the case of linear inverse problem $\bgb=\Hb\rfb+\epsilonb$, Gaussian likelihooh $p(\bgb|\rfb)=\Nc(\bgb|\Hb\rfb,\bve\Ib)$, and Gaussian prior 
$p(\rfb)=\Nc(\rfb|\bar{\fb},\bvf\Ib)$, the posterior $p(\rfb|\bgb)$ is also Gaussian: 
\beq
p(\rfb|\bgb)=\Nc(\rfb|\rfbh,\rSigmabh)
\eeq
with
\beq
\barr{l}
\rfbh=[\Hb'\Hb+\lambda\Ib]^{-1}\Hb'(\bgb-\Hb\bar{\fb})\\\
\rSigmabh=\bve[\Hb'\Hb+\lambda\Ib]^{-1}, \quad \lambda=\frac{\bve}{\bvf}
\earr
\eeq

\item 
Third step is to design an appropriate neural network which, takes as input $\bgb$ and as output $\rfbnn$. This NN can be considered as a proxy, surrogate or approximate inversion, such that its output $\rfbnn$ can be very near to the true or ground truth $\rfb$. 

\item 
Fourth step is training the NN. We distinguish two cases: 
\bit
\item Supervised case where we have a set of outputs-inputs $\{\bgb_{Ti}, \bfb_{Ti}\}$, and 
\item Unsupervised case where the available data are only $\{\bgb_{Ti}\}$. 
\eit
The parameters of the NN, $\rwb$ are estimated at the end of training step. 

\item 
Fifth step is using the trained NN, giving it any other input $\bgb_j$ which produces the outputs $\rfbh_j$ which we hope to be not very far from the ground truth $\rfb_j$. We may also want to be able to quantify the its uncertainty. 
\eit
To go more in details, first we consider the case of supervised case, and then the unsupervised case. 

\subsection{Supervized training data}

First we consider the supervised case, i.e., when we have a set of labeled  training data, (output-input of the generating forward system) : 
$\{\bgb_{Ti}, \bfb_{Ti}\}$. 
\begin{align*}
\btabu{@{}c@{}} Data \\ $\{\bgb_{Ti}, \bfb_{Ti}\}$\etabu
\Ra\fbox{~NN ($\rwb$)~}\Ra\rfb_{NNi}\Ra\fbox{~$\Hb$~}\Ra\rgbh_{NNi} \\ 
\end{align*}
To write, step by step, the relations and equations, we follow: 
\bit
\item We assume that, given the true $\rfb$ and the forward model 
$\Hb$, the the data $\{\bgb_{Ti}\}$ are generated independently:
\beq
p(\{\bgb_{Ti}\}|\{\rfb_i\})=\prod_i p(\bgb_{Ti}|\rfb_i),
\eeq 
and we can assign the likelihood $p(\bgb_{Ti}|\rfb_i)$, for example, the Gaussian:  
\beq
p(\bgb_{Ti}|\rfb_i,\bve_i\Ib) = \Nc(\bgb_{Ti}|\Hb\rfb_i,\bve_i\Ib).
\eeq
The same for $\{\bfb_{Ti}\}$, we assume that, given the true $\rfb$, we have:
\beq
p(\{\bfb_{Ti}\}|\{\rfb_i\})=\prod_i p(\bfb_{Ti}|\rfb_i)=\prod_i \Nc(\bfb_{Ti}|\rfb_i,\bvf_i\Ib)
\eeq 

\item We also assign a prior $p(\rfb)$ to the true $\rfb$ to translate our prior knowledge about it. 

\item Using the Bayes rule we get: 
\beq
p(\rfb_i|\{\bgb_{Ti}, \bfb_{Ti}\}) \propto \prod_i p(\bgb_{Ti}|\rfb_i,\bvf_i\Ib) \, p(\bfb_{Ti}|\rfb_i,\bvf_i\Ib) \, p(\rfb_i) 
\eeq

\item With the Gaussian cases of the equations [5,6], we can rewrite the posterior as: 
\begin{align}
p(\rfb_i|\{\bgb_{Ti},\bfb_{Ti}\}) &\propto \expf{-J(\rfb_i)} \nonumber \\ 
\mbox{with:~~} &
J(\rfb_i)=
\sum_i\frac{1}{2\bve_i}\|\bgb_{Ti}-\Hb\rfb_i\|^2 +\frac{1}{2\bvf_i}\|\bfb_{Ti}-\rfb_i\|^2 + \ln p(\rfb_i)
\end{align}
\item Choosing also a Gaussian prior: $p(\rfb_i)=\Nc(\rfb_i|\bar{\bfb},\bvi\Ib)$, we get: 
\beq
J(\rfb_i)=
\sum_i 
\frac{1}{2\bvf_i} \|\bfb_{Ti}-\rfb_i\|^2 + 
\frac{1}{2\bve_i}\|\bgb_{Ti}-\Hb\rfb_i\|^2 + 
\frac{1}{2\bvi} \|\rfb_i-\bar{\bfb}\|^2
\eeq
It is then easy to see that the $p(\rfb_i|\{\bgb_{Ti},\bfb_{Ti}\})$ is Gaussian: 
\beq
p(\rfb_i|\{\bgb_{Ti},\bfb_{Ti}\})=\Nc(\rfb_i|\rfbh_i,\rSigmabh_i)
\eeq
with
\[
\barr{l}
\rfbh_i=\sum_i [\Hb'\Hb+\lambda_i\Ib]^{-1}\Hb'(\bgb_{Ti}-\Hb\bfb_{Ti}-\Hb\bar{\rfb})\\\
\rSigmabh_i=\sum_i\bve_i[\Hb'\Hb+\lambda_i\Ib]^{-1}, \quad 
\lambda_i=\frac{\bve_i}{\bvf_i}
\earr
\]
\item Now, we consider the training step of the NN where we note the output of the neural network $\rfb_{NNi}$ which is a function of NN parameters $\rwb$ and the input data $\{\bgb_{Ti}\}$, we can define the following optimization criterion: 
\beq
J(\rwb)=
\sum_i \frac{1}{\bvf_i} \|\bfb_{Ti}-\rfb_{NNi}(\rwb)\|^2 + \frac{1}{\bve_i}\|\bgb_{Ti}-\Hb\rfb_{NNi}(\rwb)\|^2 + \frac{1}{\bve_i}\|\bar{\bfb} - \rfb_{NNi}(\rwb)\|^2 
\eeq
This criterion can be considered as a physics based or physics informed neural network (PINN), 
where the classical output residual part is: 
\beq
J_{NN}(\rwb)= \sum_i\frac{1}{2\bve_i} \|\bfb_{Ti}-\rfb_{NNi}(\rwb)\|^2
\eeq 
and the physics informed part is:
\beq
J_{PI}(\rwb)=\frac{1}{2\bvf_i} \|\bgb_{Ti}-\Hb\rfb_{NNi}(\rwb)\|^2+ \frac{1}{2\bvi}\|\bar{\bfb} - \rfb_{NNi}(\rwb)\|^2 
\eeq
\item We can also consider 
\beq
p(\rwb|\{\bgb_{Ti},\bfb_{Ti}\}) \propto \expf{-J(\rwb) - \ln p(\rwb)}
\eeq
and use it as the posterior probability distribution of the NN's parameter $\rwb$ given the training data $\{\bgb_{Ti},\bfb_{Ti}\}$. A specific choice for the prior $p(\rwb)$ can be one of the sparsity enforcing priors, such as 
\beq
p(\rwb) \propto \expf{-\gamma \|\rwb\|_\beta^\beta}
\eeq
However, in practice, the full expression of this posterior is very difficult to obtain, and even its optimization can be more difficult if the NN contains nonlinear activation function. 
\eit

\subsection{Unsupervised training data}
In the unsupervised case, we only have have a set of training data: $\{\bgb_{Ti}\}$. The classical NN methods can not be applied as there is not any reference data. The main advantage of the PINN is exactly in the fact that, it can be applied in this case. The schematic for this case is almost the same, but we have only for training data $\{\bgb_{Ti}\}$: 
\begin{align*}
\btabu{@{}c@{}} Data \\ $\{\bgb_{Ti}\}$\etabu
\Ra\fbox{~NN ($\rwb$)~}\Ra\rfb_{NNi}\Ra\fbox{~$\Hb$~}\Ra\rgbh_{NNi} \\ 
\end{align*}
However, we can still follow the same steps we had for the supervised case:  
\bit
\item We assume that, given the true $\rfb$ and the forward model $\Hb$, the the data $\{\bgb_{Ti}\}$ are independent, and follow a Gaussian distribution: 
\beq
p(\{\bgb_{Ti}\}|\rfb_i) = \sum_i \Nc(\bgb_{Ti}|\Hb\rfb_i,\bve_i\Ib)\propto\expf{-\frac{1}{2\bve_i}\|\bgb_{Ti}-\Hb\rfb_i\|^2}
\eeq
\item We also assign a prior $p(\rfb)$ to the true $\rfb$ to translate our prior knowledge about it. This time, we may use a more informative prior, such as Tikhonov, Total Variation or any other sparsity enforcing prior 
\beq
p(\rfb_i)\propto \expf{-\gamma\|\Db \rfb_i\|_\beta^\beta}
\eeq
\item Using the Bayes rule we get: 
\begin{align}
p(\rfb_i|\{\bgb_{Ti}\}) &\propto \expf{-J(\rfb_i)} \nonumber\\ 
\mbox{with:~~} 
J(\rfb_i)&=\sum_i \frac{1}{\bve_i}\|\bgb_{Ti}-\Hb\rfb_i\|^2 +
\gamma\|\Db\rfb_{Ti}\|_\beta^\beta
\end{align}
\item Again, as in the supervised case, we can consider 
\begin{align}
p(\rwb|\{\bgb_{Ti}\}) &\propto \expf{-J(\rwb)} \nonumber\\ 
\mbox{with:~~} 
J(\rwb)&=\sum_i \frac{1}{\bve_i}\|\bgb_{Ti}-\Hb\rfb_i(\rwb)\|^2 +
\gamma\|\Db\rfb_{Ti}(\rwb)\|_\beta^\beta +
\gamma_w\|\rwb\|_{\beta_w}^{\beta_w}
\end{align}
which can be used for the estimation of the NN's parameters. 
\eit

In both supervised and unsupervised cases, we still have to choose an appropriate structure for the NN, its depth, its number of hidden variable, as well as choosing appropriate optimization algorithms, it learning rate, etc. These difficulties make the implementation of such methods not very easy and many experience are needed to implement, to train and to use them in real applications. 
For a discussion on the choice of the structure of the NN refers to \cite{psf2023009014,proceedings2019033016,} 

\section{Conclusions}
As discussed, inverse problems arise in many science and engineering applications where we need to infer on a quantity from indirect observation. The main application cases are medical, biomedical, geophysics, and industrial non destructive testing and diagnostics using imaging systems.

In this work, a new Bayesian framework for the concept of PINN (BPINN) is presented and discussed which includes the deterministic one if we use the Maximum A Posteriori (MAP) estimation framework. 

We considered two cases of supervised and unsupervised training steps and drived, first the expressions of the posterior of the unknown desired variables, and then deduced the posterior law of the NN's parameters. However, the full use of these expressions in real applications still need a great number of challenges: the choice of the structure of the NN, its effective implementation, choice of parameters of the optimization criteria, as well as all the parameters during the learning step of the NN. 

\section*{References}
\bibliographystyle{plain}
\bibliography{refs,references,entropy-v08-i02_20250218}

\end{document}

%% file: alphabet1.tex
%


\typeout{\space}
\typeout{\space\space\space\space Fichier 'alphabet.tex' -- LSS 1994/95/99}
\typeout{\space\space\space\space (necessite les packages xspace, amsmath, bbm)}
\typeout{\space}



\def\XS{\xspace}
\DeclareMathAlphabet{\mathb}{OML}{cmm}{b}{it}
\def\sbmm#1{\ensuremath{\boldsymbol{#1}}}          
\def\bm#1{\mbox{\boldmath $#1$}}

\def\Ab{{\sbm{A}}\XS}  \def\ab{{\sbm{a}}\XS}
\def\Bb{{\sbm{B}}\XS}  \def\bb{{\sbm{b}}\XS}
\def\Cb{{\sbm{C}}\XS}  \def\cb{{\sbm{c}}\XS}
\def\Db{{\sbm{D}}\XS}  \def\db{{\sbm{d}}\XS}
\def\Eb{{\sbm{E}}\XS}  \def\eb{{\sbm{e}}\XS}
\def\Fb{{\sbm{F}}\XS}  \def\fb{{\sbm{f}}\XS}
\def\Gb{{\sbm{G}}\XS}  \def\gb{{\sbm{g}}\XS}
\def\Hb{{\sbm{H}}\XS}  \def\hb{{\sbm{h}}\XS}
\def\Ib{{\sbm{I}}\XS}  \def\ib{{\sbm{i}}\XS}
\def\Jb{{\sbm{J}}\XS}  \def\jb{{\sbm{j}}\XS}
\def\Kb{{\sbm{K}}\XS}  \def\kb{{\sbm{k}}\XS}
\def\Lb{{\sbm{L}}\XS}  \def\lb{{\sbm{l}}\XS}
\def\Mb{{\sbm{M}}\XS}  \def\mb{{\sbm{m}}\XS}
\def\Nb{{\sbm{N}}\XS}  \def\nb{{\sbm{n}}\XS}
\def\Ob{{\sbm{O}}\XS}  \def\ob{{\sbm{o}}\XS}
\def\Pb{{\sbm{P}}\XS}  \def\pb{{\sbm{p}}\XS}
\def\Qb{{\sbm{Q}}\XS}  \def\qb{{\sbm{q}}\XS}
\def\Rb{{\sbm{R}}\XS}  \def\rb{{\sbm{r}}\XS}
\def\Sb{{\sbm{S}}\XS}  \def\sb{{\sbm{s}}\XS}
\def\Tb{{\sbm{T}}\XS}  \def\tb{{\sbm{t}}\XS}
\def\Ub{{\sbm{U}}\XS}  \def\ub{{\sbm{u}}\XS}
\def\Vb{{\sbm{V}}\XS}  \def\vb{{\sbm{v}}\XS}
\def\Wb{{\sbm{W}}\XS}  \def\wb{{\sbm{w}}\XS}
\def\Xb{{\sbm{X}}\XS}  \def\xb{{\sbm{x}}\XS}
\def\Yb{{\sbm{Y}}\XS}  \def\yb{{\sbm{y}}\XS}
\def\Zb{{\sbm{Z}}\XS}  \def\zb{{\sbm{z}}\XS}

\def\Ac{{\scu{A}}\XS}   
\def\Bc{{\scu{B}}\XS}   
\def\Cc{{\scu{C}}\XS}   
\def\Dc{{\scu{D}}\XS}   
\def\Ec{{\scu{E}}\XS}   
\def\Fc{{\scu{F}}\XS}   
\def\Gc{{\scu{G}}\XS}   
\def\Hc{{\scu{H}}\XS}   
\def\Ic{{\scu{I}}\XS}   
\def\Jc{{\scu{J}}\XS}   
\def\Kc{{\scu{K}}\XS}   
\def\Lc{{\scu{L}}\XS}   
\def\Mc{{\scu{M}}\XS}   
\def\Nc{{\scu{N}}\XS}   
\def\Oc{{\scu{O}}\XS}   
\def\Pc{{\scu{P}}\XS}   
\def\Qc{{\scu{Q}}\XS}   
\def\Rc{{\scu{R}}\XS}   
\def\Sc{{\scu{S}}\XS}   
\def\Tc{{\scu{T}}\XS}   
\def\Uc{{\scu{U}}\XS}   
\def\Vc{{\scu{V}}\XS}   
\def\Wc{{\scu{W}}\XS}   
\def\Xc{{\scu{X}}\XS}   
\def\Yc{{\scu{Y}}\XS}   
\def\Zc{{\scu{Z}}\XS}   



  \def\rv{{\sbv{r}}\XS}

\def\alphab{{\sbmm{\alpha}}\XS}
\def\betab{{\sbmm{\beta}}\XS}
\def\gammab{{\sbmm{\gamma}}\XS}      
\def\deltab{{\sbmm{\delta}}\XS}      \def\Deltab{{\sbmm{\Delta}}\XS}
\def\epsilonb{{\sbmm{\epsilon}}\XS}

\def\etab{{\sbmm{\eta}}\XS}
\def\thetab{{\sbmm{\theta}}\XS}      \def\Thetab{{\sbmm{\Theta}}\XS}

\def\lambdab{{\sbmm{\lambda}}\XS}     \def\Lambdab{{\sbmm{\Lambda}}\XS}
\def\mub{{\sbmm{\mu}}\XS}
\def\nub{{\sbmm{\nu}}\XS}
\def\xib{{\sbmm{\xi}}\XS}

\def\rhob{{\sbmm{\rho}}\XS}

\def\sigmab{{\sbmm{\sigma}}\XS}      	\def\Sigmab{{\sbmm{\Sigma}}\XS}
   
\def\phib{{\sbmm{\phi}}\XS}        		\def\Phib{{\sbmm{\Phi}}\XS}

\def\chib{{\sbmm{\chi}}\XS}
\def\psib{{\sbmm{\psi}}\XS}        		\def\Psib{{\sbmm{\Psi}}\XS}
\def\omegab{{\sbmm{\omega}}\XS}      	
\def\taub{{\sbmm{\tau}}\XS}
   	
\def\zetab{{\sbmm{\zeta}}\XS}


  \def\fh{{\widehat{f}}\XS}

\def\Mh{{\widehat{M}}\XS}

  \def\qh{{\widehat{q}}\XS}
  \def\rh{{\widehat{r}}\XS}
  
  \def\th{{\widehat{t}}\XS}
  
  \def\vh{{\widehat{v}}\XS}
  \def\wh{{\widehat{w}}\XS}
  \def\xh{{\widehat{x}}\XS}
  
  \def\zh{{\widehat{z}}\XS}

  \def\qt{{\widetilde{q}}\XS}
  \def\zt{{\widetilde{z}}\XS}

\def\Abt{{\widetilde{\Ab}}\XS}  \def\abt{{\widetilde{\ab}}\XS}

\def\Dbt{{\widetilde{\Db}}\XS}  
  
  \def\fbt{{\widetilde{\fb}}\XS}

  \def\kbt{{\widetilde{\kb}}\XS}

\def\Zbt{{\widetilde{\Zb}}\XS}  \def\zbt{{\widetilde{\zb}}\XS}

\def\Abh{{\widehat{\Ab}}\XS}  \def\abh{{\widehat{\ab}}\XS}
\def\Bbh{{\widehat{\Bb}}\XS}  
  
  \def\dbh{{\widehat{\db}}\XS}
  
\def\Fbh{{\widehat{\Fb}}\XS}  \def\fbh{{\widehat{\fb}}\XS}
  \def\gbh{{\widehat{\gb}}\XS}
  \def\hbh{{\widehat{\hb}}\XS}

\def\Pbh{{\widehat{\Pb}}\XS}  
  \def\qbh{{\widehat{\qb}}\XS}
  
  \def\sbh{{\widehat{\sb}}\XS}
  \def\tbh{{\widehat{\tb}}\XS}
  \def\ubh{{\widehat{\ub}}\XS}
\def\Vbh{{\widehat{\Vb}}\XS}  \def\vbh{{\widehat{\vb}}\XS}
  
  \def\xbh{{\widehat{\xb}}\XS}
  \def\ybh{{\widehat{\yb}}\XS}
  \def\zbh{{\widehat{\zb}}\XS}

\def\alphah{{\widehat{\alpha}}\XS}
\def\betah{{\widehat{\beta}}\XS}

\def\thetah{{\widehat{\theta}}\XS}

\def\lambdah{{\widehat{\lambda}}\XS}     
\def\muh{{\widehat{\mu}}\XS}

\def\rhoh{{\widehat{\rho}}\XS}

\def\sigmah{{\widehat{\sigma}}\XS}

\def\tauh{{\widehat{\tau}}\XS}

\def\alphat{{\widetilde{\alpha}}\XS}
\def\betat{{\widetilde{\beta}}\XS}
\def\gammat{{\widetilde{\gamma}}\XS}

\def\etat{{\widetilde{\eta}}\XS}
\def\thetat{{\widetilde{\theta}}\XS}

\def\mut{{\widetilde{\mu}}\XS}

      	\def\Sigmat{{\widetilde{\Sigma}}\XS}

\def\taut{{\widetilde{\tau}}\XS}
   	
\def\zetat{{\widetilde{\zeta}}\XS}

\def\alphabh{{\widehat{\alphab}}\XS}
\def\betabh{{\widehat{\betab}}\XS}

\def\etabh{{\widehat{\etab}}\XS}
\def\thetabh{{\widehat{\thetab}}\XS}      \def\Thetabh{{\widehat{\Thetab}}\XS}

\def\lambdabh{{\widehat{\lambdab}}\XS}     
\def\mubh{{\widehat{\mub}}\XS}

\def\rhobh{{\widehat{\rhob}}\XS}

\def\sigmabh{{\widehat{\sigmab}}\XS}      	\def\Sigmabh{{\widehat{\Sigmab}}\XS}


\def\thetabt{{\widetilde{\thetab}}\XS}

\def\mubt{{\widetilde{\mub}}\XS}
\def\nubt{{\widetilde{\nub}}\XS}

      	\def\Sigmabt{{\widetilde{\Sigmab}}\XS}

%% file: alphabet2.tex

\def\blue#1{{\color{blue}#1}}
\def\red#1{{\color{red}#1}} 
\def\green#1{{\color{green}#1}} 

\def\rF{{\red{F}}}  
\def\rFb{{\red{\Fb}}}  \def\rfj{{\red{f_j}}}
\def\rZb{{\red{\Zb}}}  \def\rzj{{\red{z_j}}}

\def\rtaub{{\red{\taub}}}

\def\rvzj{\red{{v_z}_j}}

\def\bG{{\blue{G}}}  \def\bg{{\blue{g}}}

\def\rAb{{\red{\Ab}}}  \def\rab{{\red{\ab}}}
\def\rBb{{\red{\Bb}}}  \def\rbb{{\red{\bb}}}

\def\rFb{{\red{\Fb}}}  \def\rfb{{\red{\fb}}}
  \def\rgb{{\red{\gb}}}
\def\rHb{{\red{\Hb}}}  \def\rhb{{\red{\hb}}}

  \def\rmb{{\red{\mb}}}

  \def\rqb{{\red{\qb}}}

  \def\rtb{{\red{\tb}}}
  \def\rub{{\red{\ub}}}
\def\rVb{{\red{\Vb}}}  \def\rvb{{\red{\vb}}}
  \def\rwb{{\red{\wb}}}

\def\rZb{{\red{\Zb}}}  \def\rzb{{\red{\zb}}}


\def\bFb{{\blue{\Fb}}}  \def\bfb{{\blue{\fb}}}
  \def\bgb{{\blue{\gb}}}
\def\bHb{{\blue{\Hb}}}

\def\rAbh{{\widehat{\rAb}}}  \def\rabh{{\widehat{\rab}}}
\def\rBbh{{\widehat{\rBb}}}

\def\rFbh{{\widehat{\rFb}}}  \def\rfbh{{\widehat{\rfb}}}
  \def\rgbh{{\widehat{\rgb}}}
  \def\rhbh{{\widehat{\rhb}}}

  \def\rmbh{{\widehat{\rmb}}}

\def\rPbh{{\widehat{\rPb}}}  
  \def\rqbh{{\widehat{\rqb}}}

  \def\rubh{{\widehat{\rub}}}
\def\rVbh{{\widehat{\rVb}}}  \def\rvbh{{\widehat{\rvb}}}

  \def\rzbh{{\widehat{\rzb}}}


  \def\rfbt{{\widetilde{\rfb}}}

\def\rVbt{{\widetilde{\rVb}}}  \def\rvbt{{\widetilde{\rvb}}}

  \def\rzbt{{\widetilde{\rzb}}}

\def\ralphab{{\red{\alphab}}}
\def\rbetab{{\red{\betab}}}

\def\retab{{\red{\etab}}}
\def\rthetab{{\red{\thetab}}}

\def\rlambdab{{\red{\lambdab}}}     

\def\rnub{{\red{\nub}}}

      	\def\rSigmab{{\red{\Sigmab}}}
   
\def\rphib{{\red{\phib}}}

\def\rtaub{{\red{\taub}}}

\def\ralphabh{{\red{\alphabh}}}

\def\rthetabh{{\red{\thetabh}}}

      	\def\rSigmabh{{\red{\Sigmabh}}}

\def\rtaubh{{\red{\taubh}}}

\def\ralphabt{{\red{\alphabt}}}

\def\retabt{{\red{\etabt}}}
\def\rthetabt{{\red{\thetabt}}}

      	\def\rSigmabt{{\red{\Sigmabt}}}


%% file: macros_gpi.tex
\def\bdoc{\begin{document}}             \def\edoc{\end{document}}
\def\babs{\begin{abstract}}             \def\eabs{\end{abstract}}
\def\bcc{\begin{center}}                \def\ecc{\end{center}}
\def\ben{\begin{enumerate}}             \def\een{\end{enumerate}}
\def\bit{\begin{itemize}}               \def\eit{\end{itemize}}
\def\bpic{\begin{picture}}              \def\epic{\end{picture}}
\def\barr{\begin{array}}                \def\earr{\end{array}}
\def\btabu{\begin{tabular}}             \def\etabu{\end{tabular}}
\def\bcc{\begin{center}}                \def\ecc{\end{center}}
\def\bfig{\begin{figure}}               \def\efig{\end{figure}}
\def\bver{\begin{verb}}					\def\ever{\end{verb}}
\def\bbib{\begin{thebibliography}{999}}	\def\ebib{\end{thebibliography}}

\def\beq{\begin{equation}}				\def\eeq{\end{equation}}
\def\bmult{\begin{multline}}			\def\emult{\end{multline}}
\def\balign{\begin{align}}				\def\ealign{\end{align}}
\def\beqn{\begin{eqnarray}}             \def\eeqn{\end{eqnarray}}
\def\beqnx{\begin{eqnarray*}}           \def\eeqnx{\end{eqnarray*}}

\def\bm#1{\mbox{\boldmath $#1$}}

\def\zerob{{\bm 0}}
\def\oneb{{\bm 1}}

\def\ab{{\bm a}}
\def\bb{{\bm b}}
\def\cb{{\bm c}}
\def\db{{\bm d}}
\def\eb{{\bm e}}
\def\fb{{\bm f}}
\def\gb{{\bm g}}
\def\hb{{\bm h}}
\def\ib{{\bm i}}
\def\jb{{\bm j}}
\def\kb{{\bm k}}
\def\lb{{\bm l}}
\def\mb{{\bm m}}
\def\nb{{\bm n}}
\def\ob{{\bm o}}
\def\pb{{\bm p}}
\def\qb{{\bm q}}
\def\rb{{\bm r}}
\def\sb{{\bm s}}
\def\tb{{\bm t}}
\def\ub{{\bm u}}
\def\vb{{\bm v}}
\def\wb{{\bm w}}
\def\xb{{\bm x}}
\def\yb{{\bm y}}
\def\zb{{\bm z}}

\def\Ab{{\bm A}}
\def\Bb{{\bm B}}
\def\Cb{{\bm C}}
\def\Db{{\bm D}}
\def\Eb{{\bm E}}
\def\Fb{{\bm F}}
\def\Gb{{\bm G}}
\def\Hb{{\bm H}}
\def\Ib{{\bm I}}
\def\Jb{{\bm J}}
\def\Kb{{\bm K}}
\def\Lb{{\bm L}}
\def\Mb{{\bm M}}
\def\Nb{{\bm N}}
\def\Ob{{\bm O}}
\def\Pb{{\bm P}}
\def\Qb{{\bm Q}}
\def\Rb{{\bm R}}
\def\Sb{{\bm S}}
\def\Tb{{\bm T}}
\def\Ub{{\bm U}}
\def\Vb{{\bm V}}
\def\Wb{{\bm W}}
\def\Xb{{\bm X}}
\def\Yb{{\bm Y}}
\def\Zb{{\bm Z}}

\def\alphab{\bm{\alpha}}
\def\betab{\bm{\beta}}
\def\deltab{\bm{\delta}}
\def\etab{\bm{\eta}}
\def\epsilonb{\bm{\epsilon}}
\def\gammab{\bm{\gamma}}
\def\omegab{\bm{\omega}}
\def\etab{\bm{\eta}}
\def\thetab{\bm{\theta}}
\def\xib{\bm{\xi}}
\def\lambdab{\bm{\lambda}}
\def\taub{\bm{\tau}}
\def\phib{\bm{\phi}}
\def\mub{\bm{\mu}}
\def\psib{\bm{\psi}}
\def\chib{\bm{\chi}}
\def\sigmab{\bm{\sigma}}

\def\Deltab{\bm{\Delta}}
\def\Lambdab{\bm{\Lambda}}
\def\Phib{\bm{\Phi}}
\def\Psib{\bm{\Psi}}
\def\Sigmab{\bm{\Sigma}}

\def\Ac{{\cal A}}
\def\Bc{{\cal B}}
\def\Cc{{\cal C}}
\def\Dc{{\cal D}}
\def\Ec{{\cal E}}
\def\Fc{{\cal F}}
\def\Gc{{\cal G}}
\def\Hc{{\cal H}}
\def\Ic{{\cal I}}
\def\Jc{{\cal J}}
\def\Kc{{\cal K}}
\def\Lc{{\cal L}}
\def\Mc{{\cal M}}
\def\Nc{{\cal N}}
\def\Oc{{\cal O}}
\def\Pc{{\cal P}}
\def\Qc{{\cal Q}}
\def\Rc{{\cal R}}
\def\Sc{{\cal S}}
\def\Tc{{\cal T}}
\def\Uc{{\cal U}}
\def\Vc{{\cal V}}
\def\Wc{{\cal W}}
\def\Xc{{\cal X}}
\def\Yc{{\cal Y}}
\def\Zc{{\cal Z}}

\def\disp#1{\displaystyle{#1}}

\def\ra{\rightarrow}
\def\la{\leftarrow}
\def\da{\downarrow}
\def\ua{\uparrow}

\def\Ra{\Rightarrow}
\def\La{\Leftarrow}
\def\Da{\Downarrow}
\def\Ua{\Uparrow}

\def\lra{\longrightarrow}
\def\lla{\longleftarrow}
\def\Lra{\Longrightarrow}
\def\Lla{\longleftarrow}

\def\lrarr{\leftrightarrow}
\def\Lrarr{\Leftrightarrow}
\def\udarr{\updownarrow}
\def\Uparr{\Updownarrow}

\def\d#1{\,\mbox{d} #1}
\def\iii{\int_{-\infty}^{+\infty}}
\def\izi{\int_{0}^{\infty}}
\def\izpi{\int_{0}^{\pi}}
\def\izdpi{\int_{0}^{2\pi}}
\def\intd{\int\kern-.8em\int}
\def\intt{\int\kern-.8em\int\kern-.8em\int}
\def\intg{\int\kern-1.1em\int}

\def\wh{\widehat}
\def\xh{\widehat{x}}
\def\thetah{\widehat{\theta}}
\def\betah{\widehat{\beta}}

\def\xbh{\widehat{\xb}}
\def\ybh{\widehat{\yb}}
\def\thetabh{\widehat{\thetab}}
\def\etabh{\widehat{\etab}}
\def\betabh{\widehat{\betab}}

\def\xbhk{\widehat{\xb}^{k}}
\def\thetahk{\widehat{\theta}^{k}}
\def\betahk{\widehat{\beta}^{k}}

\def\xbhkp{\widehat{\xb}^{k+1}}
\def\thetahkp{\widehat{\theta}^{k+1}}
\def\betahkp{\widehat{\beta}^{k+1}}

\def\thetabhk{\widehat{\thetab}^{k}}
\def\betabhk{\widehat{\betab}^{k}}

\def\thetabhkp{\widehat{\thetab}^{k+1}}
\def\betabhkp{\widehat{\betab}^{k+1}}

\def\thetamin{\theta_{\mbox{\tiny min}}}
\def\thetamax{\theta_{\mbox{\tiny max}}}
\def\betamin{\beta_{\mbox{\tiny min}}}
\def\betamax{\beta_{\mbox{\tiny max}}}

\def\expf#1{\exp\left[ {#1} \right]}

\def\argmax#1#2{\arg\max_{#1}\left\{ #2 \right\}}
\def\argmin#1#2{\arg\min_{#1}\left\{ #2 \right\}}

\def\Prob#1{\mbox{Pr}\left\{#1\right\}}
\def\var#1{\mbox{Var}\left\{#1\right\}}
\def\cov#1{\mbox{Cov}\left\{#1\right\}}
\def\corr#1{\mbox{Corr}\left\{#1\right\}}
\def\rang#1{\mbox{rang}\left\{#1\right\}}
\def\det#1{\mbox{d\'et}\left\{#1\right\}}

\def\gbu{\underline{\gb}}
\def\fbu{\underline{\fb}}
\def\zbu{\underline{\zb}}
\def\epsilonbu{\underline{\epsilonb}}
\def\thetabu{\underline{\thetab}}

\def\ve{{v_{\epsilon}}}
\def\sigmae{{\sigma_{\epsilon}}}
\def\Sigmae{{\Sigmab_{\epsilonb}}}
\def\Sigmaf{{\Sigmab_{\fb}}}
\def\Sigmag{{\Sigmab_{\gb}}}

\def\fbuh{\widehat{\fbu}}
\def\zbuh{\widehat{\fbu}}
\def\thetabuh{\widehat{\thetabu}}

\def\fbh{\widehat{\fb}}
\def\Pbh{\widehat{\Pb}}
\def\Sigmabh{\widehat{\Sigmab}}
\def\Abh{\widehat{\Ab}}
\def\Bbh{\widehat{\Bb}}
\def\thetabh{\widehat{\thetab}}
\def\Rbe{\Rb_{\epsilon}}
\def\Rbeh{\widehat{\Rbe}}

\def\Rjk{{R_j}_k}
\def\fbik{{\fb_i}_k}
\def\fbjk{{\fb_j}_k}

\def\mik{{m_i}_k}
\def\sigmaik{{\sigma_i^2}_k}
\def\Sigmaik{{\Sigmab_i}_k}
\def\alphaik{{\alpha_i}_k}
\def\betaik{{\beta_i}_k}

\def\muik{{\mu_i}_k}
\def\vik{{v_i}_k}

\def\mjk{{m_j}_k}
\def\sigmajk{{\sigma_j^2}_k}
\def\Sigmajk{{\Sigmab_j}_k}
\def\alphajk{{\alpha_j}_k}
\def\betajk{{\beta_j}_k}

\def\mujk{{\mu_j}_k}
\def\vjk{{v_j}_k}
\def\njk{{n_j}_k}

\def\alphae{\alpha_{\epsilon}}
\def\betae{\beta_{\epsilon}}
\def\alphaez{\alpha_{\epsilon_0}}
\def\betaez{\beta_{\epsilon_0}}
\def\alphaei{\alpha_{\epsilon_i}}
\def\betaei{\beta_{\epsilon_i}}
\def\alphaeiz{\alpha_{\epsilon_{i_0}}}
\def\betaeiz{\beta_{\epsilon_{i-0}}}

\def\mbz{{{\mb}_z}}
\def\Sigmabz{{{\Sigmab}_z}}
\def\diag#1{\mbox{diag}\left[#1\right]}

\def\sigmah{\widehat{\sigma}}
\def\fh{\widehat{f}}
\def\sigmaz{\sigma_z}

\def\sumi{\sum_{i=1}^{M} }
\def\sumj{\sum_{j=1}^{N} }
\def\lambdah{\wh{\lambda}}
\def\lambdabh{\wh{\lambdab}}

\def\det#1{\hbox{det}\left(#1\right)}
\def\defined{\,\shortstack{$\triangle$\\ =}\,}

\def\zmat{\left[\matrix{0}\right]}
\def\det#1{\hbox{det}(#1)}
\def\th{\widehat{t}}
\def\xh{\widehat{x}}
\def\lambdah{\widehat{\lambda}}
\def\muh{\widehat{\mu}}
\def\sigmah{\widehat{\sigma}}
\def\rhoh{\widehat{\rho}}
\def\betah{\widehat{\beta}}
\def\alphah{\widehat{\alpha}}
\def\tauh{\widehat{\tau}}

\def\tbh{\widehat{\tb}}
\def\mubh{\widehat{\mub}}
\def\sigmabh{\widehat{\sigmab}}
\def\rhobh{\widehat{\rhob}}
\def\oneb{\mbox{\bf 1}}

\def\bm#1{\mbox{\boldmath $#1$}}
\def\zerob{{\bf 0}}
\def\oneb{{\bf 1}}
\def\intg{\int\kern-1.1em\int}
\def\expf#1{\exp\left[ {#1} \right]}

\def\gbu{\underline{\gb}}
\def\fbu{\underline{\fb}}
\def\zbu{\underline{\zb}}
\def\epsilonbu{\underline{\epsilonb}}
\def\uthetab{\underline{\thetab}}

\def\sigmae{{\sigma_{\epsilon}}}
\def\Sigmae{{\Sigma_{\epsilonb}}}
\def\Sigmabe{{\Sigmab_{\epsilonb}}}
\def\Sigmaf{{\Sigmab_{\fb}}}
\def\Sigmag{{\Sigmab_{\gb}}}

\def\fbuh{\widehat{\fbu}}
\def\zbuh{\widehat{\fbu}}
\def\uthetabh{\widehat{\uthetab}}

\def\fbh{\widehat{\fb}}
\def\Sigmabh{\widehat{\Sigmab}}
\def\Abh{\widehat{\Ab}}
\def\thetabh{\widehat{\thetab}}

\def\Rjk{{R_j}_k}
\def\fbik{{\fb_i}_k}
\def\fbjk{{\fb_j}_k}

\def\mik{{m_i}_k}
\def\sigmaik{{\sigma_i^2}_k}
\def\Sigmaik{{\Sigmab_i}_k}
\def\alphaik{{\alpha_i}_k}
\def\betaik{{\beta_i}_k}

\def\muik{{\mu_i}_k}
\def\vik{{v_i}_k}

\def\mjk{{m_j}_k}
\def\sigmajk{{\sigma_j^2}_k}
\def\Sigmajk{{\Sigmab_j}_k}
\def\alphajk{{\alpha_j}_k}
\def\betajk{{\beta_j}_k}

\def\mujk{{\mu_j}_k}
\def\vjk{{v_j}_k}
\def\njk{{n_j}_k}

\def\mbz{{{\mb}_z}}
\def\Sigmabz{{{\Sigmab}_z}}

\def\vect{\mbox{vect}}
\def\lra{\longrightarrow}

\def\gir{g_i(\rb)}
\def\fjr{f_j(\rb)}
\def\zjr{z_j(\rb)}
\def\epsilonir{\epsilon_i(\rb)}
\def\gbr{\gb(\rb)}
\def\fbr{\fb(\rb)}
\def\zbr{\zb(\rb)}
\def\epsilonbr{\epsilonb(\rb)}
\def\fbhr{\fbh(\rb)}
\def\Sigmabhr{\Sigmabh(\rb)}

\def\mbzr{\mb_{z(\rb)}}
\def\Sigmabzr{\Sigmab_{z(\rb)}}
\def\Sigmagbzr{{\Sigmab_g}_{z(\rb)}}

\def\mjzjr#1{{m_{#1}}_{z_{#1}(\rb)}}
\def\sigmajzjr#1{{\sigma_{#1}}_{z_{#1}(\rb)}}

\def\Beta{B}
\def\Betab{\bm{\Beta}}
\def\Betabe{\bm{\Beta}_{\epsilonb}}

\def\fh{\widehat{f}}
\def\sigmah{\widehat{\sigma}}
\def\sigmaei{\sigma_{\epsilon_i}^2}

\def\thetah{\widehat{\theta}}
\def\sigmah{\widehat{\sigma}}

\def\Ftxm{\widetilde{F}_{X}(x_-)}
\def\Ftxp{\widetilde{F}_{X}(x_+)}

\def\NU{{\cal V}}

\def\tFx{\widetilde{F}_{X}}

\def\Fx{F_{X}(x)}
\def\fx{f_{X}(x)}

\def\Fxv{F_{X}(x)}
\def\fxv{f_{X}(x)}

\def\Fxi{{F}_{X_i}(x_i)}
\def\fxi{{f}_{X_i}(x_i)}

\def\Gnu{G_{\NU}(\nu)}
\def\gnu{g_{\NU}(\nu)}

\def\Fnu{F_{\NU}(\nu)}
\def\fnu{f_{\NU}(\nu)}

\def\Finnu#1{F^{-1}_{\NU}(#1)}

\def\Gnua#1{G_{\NU}(#1)}
\def\Gnub#1{G_{\NU}^{-1}(#1)}
\def\Gnuh{G_{\NU}^{-1}(1/2)}

\def\Ftx{\widetilde{F}_{X}(x)}
\def\ftx{\widetilde{f}_{X}(x)}

\def\Ftxi{\widetilde{F}_{X_i}(x_i)}
\def\ftxi{\widetilde{f}_{X_i}(x_i)}

\def\Ftxv{\widetilde{F}_{X}(x)}
\def\FtX#1{\widetilde{F}_{X}(#1)}
\def\ftxv{\widetilde{f}_{X}(x)}

\def\fxnu{f_{X|\NU}(x|\nu)}
\def\fxNU{f_{X|\NU}(x|\NU)}
\def\FxNU{F_{X|\NU}(x|\NU)}

\def\fxNUtheta{f_{X|\NU,\theta}(x|\NU,\theta)}
\def\FxNUtheta{F_{X|\NU,\theta}(x|\NU,\theta)}

\def\Fxnu{F_{X|\NU}(x|\nu)}
\def\Fxnui#1{F_{X|\NU}(x|\nu_#1)}

\def\FxN#1{F_{X|\NU}(#1)}
\def\FxNU{F_{X|\NU}(x|\NU)}
\def\Fxnun#1{F_{X|\NU}(x|#1)}
\def\Fxinun#1{F_{X_i|\NU}(x_i|#1)}
\def\fxnu{f_{X|\NU}(x|\nu)}
\def\fxNU{f_{X|\NU}(x|\NU)}

\def\Fxvnu{F_{X|\NU}(x|\nu)}
\def\Fxvnun#1{F_{X|\NU}(x|#1)}
\def\FxvNU{F_{X|\NU}(x|\NU)}

\def\FXvnu{F_{X|\NU}}

\def\fxvnu{f_{X|\NU}(x|\nu)}
\def\fxvNU{f_{X|\NU}(x|\NU)}

\def\FXvNU#1{{F}_{X|\NU}(#1|\NU)}
\def\FXvn#1{{F}_{X|\NU}(x|#1)}

\def\Fxtheta{F_{X|\theta}(x|\theta)}
\def\fxtheta{f_{X|\theta}(x|\theta)}

\def\Fxvtheta{F_{X|\theta}(x|\theta)}
\def\fxvtheta{f_{X|\theta}(x|\theta)}

\def\Ftxtheta{\widetilde{F}_{X|\theta}(x|\theta)}
\def\ftxtheta{\widetilde{f}_{X|\theta}(x|\theta)}

\def\Ftxvtheta{\widetilde{F}_{X|\theta}(x|\theta)}
\def\ftxvtheta{\widetilde{f}_{X|\theta}(x|\theta)}

\def\Fxnutheta{F_{X|\NU,\theta}(x|\nu,\theta)}
\def\fxnutheta{f_{X|\NU,\theta}(x|\nu,\theta)}

\def\ftheta{f_{\Theta}(\theta)}
\def\fthetaxnuz{f_{\Theta|X,\nu_0}(\theta|x,\nu_0)}
\def\fthetax{f_{\Theta|X}(\theta|x)}

\def\Fxvnutheta{F_{X|\NU,\theta}(x|\nu,\theta)}
\def\FxvNUtheta{F_{X|\NU,\theta}(x|\NU,\theta)}
\def\fxvnutheta{f_{X|\NU,\theta}(x|\nu,\theta)}
\def\Ftxvnutheta{\widetilde{F}_{X|\theta}(x|\theta)}
\def\ftxvnutheta{\widetilde{f}_{X|\theta}(x|\theta)}

\def\FxvNUm{F_{X|\NU}(x_-|\NU)}
\def\FxvNUp{F_{X|\NU}(x_+|\NU)}

\def\Fxinu{F_{X_i|\NU}(x_i|\nu)}
\def\fxinu{f_{X_i|\NU}(x_i|\nu)}

\def\Ftxi{\widetilde{F}_{X_i}(x_i)}
\def\ftxi{\widetilde{f}_{X_i}(x_i)}

\def\fxitheta{f_{X_i|\theta}(x_i|\theta)}
\def\ftxitheta{\widetilde{f}_{X_i|\theta}(x_i|\theta)}

\def\thetah{\widehat{\theta}}
\def\thetat{\widetilde{\theta}}

\def\lnuxv{l_{\nu|X}(\nu|x)}

\def\Lnuxva#1{L_{\nu|X}(#1 | x)}
\def\lnuxva#1{l_{\nu|X}(#1 | x)}

\def\Prob#1{\mbox{P}\left(#1\right)}

\def\lthetaxvnuz{l(\theta)}

\def\lthetaxvnuz{l_{\theta|x,\nu_0}(\theta | x,\nu_0)}
\def\fxnuztheta{f_{X|\nu_0,\theta}(x | \nu_0,\theta)}
\def\fxinuztheta{f_{X|\nu_0,\theta}(x_i | \nu_0,\theta)}
\def\fxvtheta{f_{X|\theta}(x | \theta)}
\def\lthetaxv{l_{\theta|X}(\theta | x)}
\def\ltthetaxv{\widetilde{l}_{\theta|X}(\theta | x)}

\def\ltheta{l(\theta )}
\def\lttheta{\widetilde{l}(\theta)}
\def\Lnu{L(\nu )}
\def\Lnui#1{L(\nu_#1)}
\def\Linu{L_i(\nu )}
\def\Linu{L_i(\nu )}
\def\LNU{L(\NU )}
\def\Lin#1{L^{-1}(#1)}
\def\L#1{L(#1)}
\def\Li#1{L_i(#1)}
\def\FNU#1{F_\NU(#1)}
\def\FinNU#1{F^{-1}_\NU(#1)}

\def\ftthetax{\tilde{f}_{\Theta|X}(\theta|x)}
\def\ER{R}

\def\Ndd#1#2#3{{\cal N}\left( #1 ;~ #2, #3 \right)}
\def\Cdd#1#2#3{{\cal C}\left( #1 ;~ #2, #3 \right)}
\def\Sdd#1#2#3#4{{\cal S}\left( #1 ;~ #2, #3, #4 \right)}
\def\Edd#1#2{{\cal E}\left( #1 ;~ #2 \right)}
\def\DEdd#1#2{{\cal DE}\left( #1 ;~ #2 \right)}
\def\Gdd#1#2#3{{\cal G}\left( #1 ;~ #2, #3 \right)}
\def\IGdd#1#2#3{{\cal IG}\left( #1 ;~ #2, #3 \right)}

\def\Xwr{X(\omega,\rb)}
\def\xwr{x(\omega,\rb)}

\def\muzw{\mu_{z}(\omega)}
\def\muzwr{\mu_{z(\rb)}(\omega)}
\def\szw{\sigma_{z}^2(\omega)}
\def\szwr{\sigma_{z(\rb)}^2(\omega)}
\def\pzw{p_{z}(\omega)}
\def\pzwr{p_{z(\rb)}(\omega)}

\def\hszw{\widehat{\sigma_{z}^2}(\omega)}
\def\hmuzw{\widehat{\mu}_{z}(\omega)}
\def\hpzw{\widehat{p}_{z}(\omega)}

\def\muzwrl{\mu_{z(\rb)}(\omega-l)}

\def\pxwrcz{p\left(\xwr ~|~ z\right)}
\def\pxwr{p\left(\xwr\right)}

\def\Xbw{\Xb(\omega)}
\def\xbw{\xb(\omega)}
\def\uXb{\underline{\Xb}}
\def\uxb{\underline{\xb}}
\def\pxbw{p(\xbw)}
\def\puxb{p(\uxb)}
\def\pzb{P(\zb)}

\def\Szw{S_z(\omega)}
\def\mubw{\mub(\omega)}

\def\Srw{S_{\rb}(\omega)}
\def\Swr{S_{\omega}(\rb)}
\def\Sbw{\Sb(\omega)}
\def\sbw{\sb(\omega)}
\def\Sbr{\Sb(\rb)}
\def\mubw{\mub(\omega)}
\def\cov#1{\mbox{cov}[#1]}

\def\pdf{\emph{pdf}}
\def\pmf{\emph{pmf}}
\def\dpdx#1#2{\frac{\partial #1}{\partial #2}}

\def\REM#1{}

\def\XXwr{X(\omega,\rb)}
\def\xxwr{x(\omega,\rb)}

\def\Xwr{X_{\omega}(\rb)}
\def\xwr{x_{\omega}(\rb)}
\def\Xrw{X_{\rb}(\omega)}
\def\xrw{x_{\rb}(\omega)}

\def\Ywr{Y_{\omega}(\rb)}
\def\ywr{y_{\omega}(\rb)}
\def\Yrw{Y_{\rb}(\omega)}
\def\yrw{y_{\rb}(\omega)}

\def\Ewr{E_{\omega}(\rb)}
\def\ewr{\epsilon_{\omega}(\rb)}
\def\Erw{E_{\rb}(\omega)}
\def\erw{\epsilon_{\rb}(\omega)}

\def\Xbr{\Xb(\rb)}
\def\xbr{\xb(\rb)}
\def\Xbw{\Xb(\omega)}
\def\xbw{\xb(\omega)}

\def\uXb{\underline{\Xb}}
\def\uxb{\underline{\xb}}
\def\uXbz{\underline{\Xb}_z}
\def\uxbz{\underline{\xb}_z}

\def\uthetab{\underline{\thetab}}
\def\uthetabz{\underline{\thetab}_z}

\def\Ywr{Y_{\omega}(\rb)}
\def\ywr{y_{\omega}(\rb)}
\def\Yrw{Y_{\rb}(\omega)}
\def\yrw{y_{\rb}(\omega)}
\def\Ybr{\Yb(\rb)}
\def\ybr{\yb(\rb)}
\def\Ybw{\Yb(\omega)}
\def\ybw{\yb(\omega)}
\def\uYb{\underline{\Yb}}
\def\uyb{\underline{\yb}}
\def\uYbz{\underline{\Yb}_z}
\def\uybz{\underline{\yb}_z}

\def\Ewr{E_{\omega}(\rb)}
\def\ewr{\epsilon_{\omega}(\rb)}
\def\Erw{E_{\rb}(\omega)}
\def\erw{\epsilon_{\rb}(\omega)}
\def\Ebr{Eb(\rb)}
\def\ebr{\epsilonb(\rb)}
\def\Ebw{Eb(\omega)}
\def\ebw{\epsilonb(\omega)}
\def\uEb{\underline{\Eb}}

\def\Erwp{E_{\rb}(\omega')}
\def\Xrwp{X_{\rb}(\omega')}

\def\xwmr{x_{\omega-1}(\rb)}
\def\muzwm{\mu_{z}(\omega-1)}

\def\rhoz{\rho_z}
\def\sez{\sigma_{\eta_z}^2}

\def\xwrp{x_{\omega}(\rb')}
\def\xrwp{x_{\rb}(\omega')}
\def\xwpr{x_{\omega'}(\rb)}
\def\xwprp{x_{\omega'}(\rb')}
\def\Xwrp{x_{\omega}(\rb')}
\def\Xwrp{X_{\omega}(\rb')}
\def\Xwpr{X_{\omega'}(\rb)}
\def\Xwprp{X_{\omega'}(\rb')}

\def\sigmaed{\sigmae^2}
\def\sigmaew{\sigmae(\omega)}
\def\sigmaedw{\sigmaed(\omega)}

\def\Sigmabe{\Sigmab_{\epsilon}}
\def\sigmabh{\widehat{\Sigmab}}
\def\xbh{\widehat{\xb}}
\def\zbh{\widehat{\zb}}
\def\qbh{\widehat{\qb}}
\def\sbh{\widehat{\sb}}

\def\espx#1#2{\mbox{E}_{#1}\left\{#2\right\}}
\def\esp#1{\mbox{E}\left\{#1\right\}}
\def\d#1{\mbox{~d}#1}

\def\Tr#1{\mbox{Tr}\left[#1\right]}

\def\pmatrix#1{\left[\barr{ccccccccccccccc} #1 \earr\right]}
\def\Bf{\bm{B}}
\def\gbh{\widehat{\gb}}
\def\dbh{\widehat{\db}}
\def\Tr#1{\mbox{Tr}\left[#1\right]}
\def\Cov#1{\mbox{Cov}\left[#1\right]}
\def\iid{i.i.d.~}

\def\cro#1{\left[#1\right]}
\def\acc#1{\left\{#1\right\}}
\def\aprio{\emph{a priori~}}
\def\apost{\emph{a posteriori~}}
\def\hbh{\widehat{\hb}}
\def\qh{\widehat{q}}
\def\fbt{\widetilde{\fb}}
\def\thetabt{\widetilde{\thetab}}

\def\mubt{\widetilde{\mub}}
\def\Sigmabt{\widetilde{\Sigmab}}
\def\qt{\widetilde{q}}
\def\vt{\widetilde{v}}
\def\zt{\widetilde{z}}
\def\Dbt{\widetilde{\Db}}
\def\alphat{\widetilde{\alpha}}
\def\betat{\widetilde{\beta}}
\def\abt{\widetilde{\ab}}
\def\ajt{\widetilde{a}_j}
\def\taut{\widetilde{\tau}}
\def\mut{\widetilde{\mu}}
\def\Zbt{\widetilde{\Zb}}
\def\Abt{\widetilde{\Ab}}
\def\zbt{\widetilde{\zb}}

\def\bloca#1#2#3{
\begin{picture}(50,50)
  \put(32,42){\makebox(0,0){#1}}
  \put(25,50){\vector(0,-1){20}}
  \put(0,10){\framebox(50,20){#2}}
  \put(0,0){\makebox(50,0){#3}}  
\end{picture}
}

\def\ugb{\underline{\gb}}
\def\fbh{\widehat{\fb}}
\def\zbh{\widehat{\zb}}
\def\qbh{\widehat{\qb}}
\def\thetabh{\widehat{\thetab}}
\def\Pbh{\widehat{\Pb}}

\def\sigmae{\sigma_{\epsilon}}
\def\REM#1{}
\def\fh{\widehat{f}}
\def\zh{\widehat{z}}
\def\disp{\displaystyle}
\def\oneb{{\boldmath{1}}}
\def\zerob{{\boldmath{0}}}
\def\d#1{\; \mbox{d} #1}
\def\expf#1{\exp\left\{#1\right\}}
\def\esp#1{\mbox{E}\left\{#1\right\}}
\def\lra{\longrightarrow}

\def\bslide#1{\begin{frame}{\Large\bf #1}}
\def\eslide{\end{frame}}

\def\inversions#1#2#3{
\begin{picture}(100,40)
  \put(0,15){\makebox(15,0){#1}}
  \put(20,15){\vector(1,0){10}}
  \put(30,7){\framebox(40,15){#2}}
  \put(70,15){\vector(1,0){10}}
  \put(85,15){\makebox(15,0){#3}}
\end{picture}
}

\def\blue#1{{\color{blue}#1}}
\def\red#1{{\color{red}#1}}
\def\green#1{{\color{green}#1}}

\def\rfb{\red{\fb}}
\def\bgb{\blue{\gb}}
\def\rfbh{\red{\widehat{\fb}}}
\def\rfbt{\red{\widetilde{\fb}}}
\def\rzbt{\red{\widetilde{\zb}}}
\def\rqb{\red{\qb}}
\def\rqbh{\red{\widehat{\qb}}}
\def\rzb{\red{\zb}}
\def\rzbh{\red{\widehat{\zb}}}
\def\rthetab{\red{\thetab}}
\def\retab{\red{\etab}}
\def\rthetabh{\red{\widehat{\thetab}}}
\def\rthetabt{\red{\widetilde{\thetab}}}
\def\retabt{\red{\widetilde{\etab}}}
\def\ralphabt{\red{\widetilde{\alphab}}}

\def\ugb{\underline{\gb}}
\def\fbh{\widehat{\fb}}
\def\zbh{\widehat{\zb}}
\def\qbh{\widehat{\qb}}
\def\thetabh{\widehat{\thetab}}
\def\Pbh{\widehat{\Pb}}

\def\UP#1{\uppercase{#1}}
\def\sca#1{\uppercase{#1}}

\def\sigmae{\sigma_{\epsilon}}
\def\REM#1{}
\def\fh{\widehat{f}}
\def\zh{\widehat{z}}
\def\disp{\displaystyle}
\def\oneb{{\boldmath{1}}}
\def\zerob{{\boldmath{0}}}
\def\d#1{\; \mbox{d} #1}
\def\expf#1{\exp\left\{#1\right\}}
\def\esp#1{\mbox{E}\left\{#1\right\}}
\def\lra{\longrightarrow}

\def\rfb{\red{\fb}}
\def\rFb{\red{\Fb}}
\def\bgb{\blue{\gb}}
\def\rfbh{\red{\widehat{\fb}}}
\def\rqb{\red{\qb}}
\def\rqbh{\red{\widehat{\qb}}}
\def\rzb{\red{\zb}}
\def\rzbh{\red{\widehat{\zb}}}
\def\rthetab{\red{\thetab}}
\def\ralphab{\red{\alphab}}
\def\rbetab{\red{\betab}}
\def\rthetabh{\red{\widehat{\thetab}}}
\def\ralphabh{\red{\widehat{\alphaab}}}
\def\rSigmab{\red{\Sigmab}}
\def\rSigmabe{\red{\Sigmabe}}

\def\rz{\red{z}}
\def\rzh{\red{\widehat{z}}}

\def\rab{\red{\ab}}
\def\rabh{\red{\widehat{\ab}}}
\def\rAb{\red{\Ab}}
\def\rAbh{\red{\widehat{\Ab}}}
\def\rBb{\red{\Bb}}
\def\rBbh{\red{\widehat{\Bb}}}
\def\rPbh{\red{\widehat{\Pb}}}
\def\rub{\red{\ub}}
\def\rubh{\red{\widehat{\ub}}}

\def\rVb{\red{\Vb}}
\def\rVbh{\widehat{\rVb}}
\def\rVbt{\widetilde{\rVb}}

\def\rvb{\red{\vb}}
\def\rvbh{\widehat{\rvb}}
\def\rvbt{\widetilde{\rvb}}

\def\bfb{\blue{\fb}}
\def\bFb{\blue{\Fb}}

\def\rfi{\red{f}_i}
\def\rfj{\red{f}_j}
\def\ruj{\red{u}_j}
\def\raj{\red{a}_j}
\def\rzj{\red{z}_j}
\def\rvj{\red{v}_j}
\def\rtau{\red{\tau}}
\def\rtauj{\red{\tau_j}}
\def\rtaub{\red{\taub}}
\def\rtauh{\widehat{\rtau}}
\def\rtaujh{\widehat{\rtauj}}
\def\rtaubh{\widehat{\rtaub}}
\def\rve{\red{v_{\epsilon}}}
\def\rv{\red{v}}

\def\rfjh{\red{\widehat{f}}_j}
\def\rujh{\red{\widehat{u}}_j}
\def\rajh{\red{\widehat{a}}_j}
\def\rzjh{\red{\widehat{z}}_j}
\def\rvjh{\red{\widehat{v}}_j}

\def\ralpha{\red{\alpha}}
\def\rbeta{\red{\beta}}
\def\rtheta{\red{\theta}}

\def\rlambda{\red{\lambda}}
\def\ralphah{\widehat{\ralpha}}
\def\rbetah{\widehat{\rbeta}}
\def\rthetah{\widehat{\rtheta}}
\def\rlambdah{\widehat{\rlambda}}

\def\ubh{\widehat{\ub}}

\def\rve{\red{v_{\epsilon}}}
\def\alphae{\alpha_{\epsilon_0}}
\def\betae{\beta_{\epsilon_0}}
\def\alphajh{\widehat{\alpha}_j}
\def\betajh{\widehat{\beta}_j}
\def\alphaeh{\widehat{\alpha}_{\epsilon_0}}
\def\betaeh{\widehat{\beta}_{\epsilon_0}}

\def\rqb{\red{\qb}}
\def\rqbh{\red{\widehat{\qb}}}
\def\rzb{\red{\zb}}
\def\rZb{\red{\Zb}}
\def\rab{\red{\ab}}
\def\rAb{\red{\Ab}}
\def\rvb{\red{\vb}}
\def\rmb{\red{\mb}}
\def\rzbh{\red{\widehat{\zb}}}
\def\rvbh{\red{\widehat{\vb}}}
\def\rmbh{\red{\widehat{\mb}}}
\def\rthetab{\red{\thetab}}
\def\rthetabh{\red{\widehat{\thetab}}}
\def\rfbh{\widehat{\rfb}}
\def\rhb{\red{\hb}}
\def\rhbh{\widehat{\rhb}}
\def\rthetai{\red{\theta}_i}
\def\rthetab{\red{\thetab}}
\def\rfjm{\red{f}_{j-1}}
\def\rfim{\red{f}_{i-1}}

\def\muh{\widehat{\mu}}
\def\vh{\widehat{v}}

\def\rSigmabh{\red{\Sigmabh}}
\def\rVbh{\red{\Vbh}}
\def\Fbh{\red{\Fb}}
\def\rFbh{\red{\Fbh}}

\def\Sigmabeh{\widehat{\Sigmabe}}
\def\rSigmabeh{\widehat{\red{\Sigmabe}}}

\def\rf{\red{f}}
\def\bg{\blue{g}}
\def\rF{\red{F}}
\def\bG{\blue{G}}

\def\sumk{\sum_{k=1}^K}
\def\suml{\sum_{l=1}^L}
\def\sumn{\sum_{n=1}^N}
\def\sumj{\sum_{j=1}^J}
\def\sumi{\sum_{i=1}^I}

\def\nub{\bm{\nu}}
\def\rak{\red{a_k}}
\def\rbk{\red{b_k}}
\def\rck{\red{c_k}}
\def\rtk{\red{t_k}}
\def\rvk{\red{v_k}}
\def\rnuk{\red{\nu_k}}
\def\rnuz{\red{\nu_0}}
\def\rphik{\red{\phi_k}}
\def\rnub{\red{\nub}}
\def\rtb{\red{\tb}}
\def\rphib{\red{\phib}}
\def\rfn{\red{f_n}}
\def\rtn{\red{t_n}}
\def\rvn{\red{v_n}}

\def\rve{\red{v_\epsilon}}
\def\rvf{\red{v_f}}
\def\rveh{\red{\widehat{v}_\epsilon}}
\def\rvfh{\red{\widehat{v}_f}}

\def\rnu{\red{\nu}}
\def\rphi{\red{\phi}}
\def\rbl{\red{b_l}}
\def\rbb{\red{\bb}}

\def\modela{\tt\setlength{\unitlength}{0.92pt}
\begin{picture}(250,106)
\thinlines    \put(-5,105){$\Nc(t|\red{t_1,v_1})$}
              \put(35,100){\vector(1,0){25}}
              \put(60,95){\framebox(37,16){\red{$a_1$}}}
			  \multiput(77,72)(0,7){3}{\circle*{4}}
              \put(-5,65){$\Nc(t|\red{t_k,v_k})$}
              \put(35,59){\vector(1,0){25}}
              \put(60,51){\framebox(37,16){\red{$a_k$}}}
              \multiput(77,32)(0,7){3}{\circle*{4}}              
              \put(-5,23){$\Nc(t|\red{t_K,v_K})$}
              \put(60,10){\framebox(37,16){\red{$a_K$}}}
              \put(35,18){\vector(1,0){25}}

              \put(98,103){\vector(1,-1){38}}
              \put(98,59){\vector(1,0){39}}
              \put(98,17){\vector(1,1){38}}

              \put(150,60){\circle{24}}
              \put(146,57){$\Large+$}
              \put(165,75){$g_{\rthetab}(t)$}
              \put(164,60){\vector(1,0){33}}
              
              \put(212,85){$\epsilon(t)$}
              \put(210,95){\vector(0,-1){20}}
              \put(211,60){\circle{24}}
              \put(207,58){$\Large+$}
              
              \put(225,66){$g(t)$}
              \put(225,60){\vector(1,0){20}}
\end{picture}
}

\def\modelb{\tt\setlength{\unitlength}{0.92pt}
\begin{picture}(250,106)
\thinlines    \put(-5,105){$\delta(t-\red{t_1})$}
              \put(0,100){\vector(1,0){37}}
              \put(38,95){\framebox(60,16){$\red{a_1} \Nc(t|0,\red{v_1})$}}
			  \multiput(77,72)(0,7){3}{\circle*{4}}
              \put(-5,65){$\delta(t-\red{t_k})$}
              \put(0,59){\vector(1,0){37}}
              \put(38,51){\framebox(60,16){$\red{a_k} \Nc(t|0,\red{v_k})$}}
              \multiput(77,32)(0,7){3}{\circle*{4}}              
              \put(-5,23){$\delta(t-\red{t_K})$}
              \put(38,10){\framebox(60,16){$\red{a_K} \Nc(t|0,\red{v_K})$}}
              \put(0,18){\vector(1,0){37}}

              \put(98,103){\vector(1,-1){38}}
              \put(98,59){\vector(1,0){39}}
              \put(98,17){\vector(1,1){38}}

              \put(150,60){\circle{24}}
              \put(146,57){$\Large+$}
              \put(165,75){$g_{\rthetab}(t)$}
              \put(164,60){\vector(1,0){33}}
              
              \put(212,85){$\epsilon(t)$}
              \put(210,95){\vector(0,-1){20}}
              \put(211,60){\circle{24}}
              \put(207,58){$\Large+$}
              
              \put(225,66){$g(t)$}
              \put(225,60){\vector(1,0){20}}
\end{picture}
}

\def\modelc{\tt\setlength{\unitlength}{0.92pt}
\begin{picture}(250,106)
\thinlines    \put(-5,105){$\delta(t-\red{t_1})$}
              \put(0,100){\vector(1,0){37}}
              \put(38,95){\framebox(40,16){$\red{a_1}$}}
			  \multiput(57,72)(0,7){3}{\circle*{4}}
              \put(-5,65){$\delta(t-\red{t_k})$}
              \put(0,59){\vector(1,0){37}}
              \put(38,51){\framebox(40,16){$\red{a_k}$}}
              \multiput(57,32)(0,7){3}{\circle*{4}}              
              \put(-5,23){$\delta(t-\red{t_K})$}
              \put(38,10){\framebox(40,16){$\red{a_K}$}}
              \put(0,18){\vector(1,0){37}}

              \put(78,103){\vector(1,-1){38}}
              \put(78,59){\vector(1,0){39}}
              \put(78,17){\vector(1,1){38}}

              \put(130,60){\circle{24}}
              \put(126,57){$\Large+$}
              \put(144,60){\vector(1,0){10}}
              \put(154,52){\framebox(50,16){$\Nc(t|0,v)$}}
              
              \put(205,75){$g_{\rthetab}(t)$}
              \put(204,60){\vector(1,0){33}}
              
              \put(252,85){$\epsilon(t)$}
              \put(250,95){\vector(0,-1){20}}
              \put(251,60){\circle{24}}
              \put(247,58){$\Large+$}
              
              \put(265,66){$g(t)$}
              \put(265,60){\vector(1,0){20}}
\end{picture}
}

\def\modeld{\tt\setlength{\unitlength}{0.92pt}
\begin{picture}(250,106)
\thinlines    \put(20,70){$s(t)=\sum_k \rak \delta(t-\red{t_k})$}
              \put(105,60){\vector(1,0){50}}
              \put(154,52){\framebox(50,16){$\Nc(t|0,v)$}}
              
              \put(205,75){$g_{\rthetab}(t)$}
              \put(204,60){\vector(1,0){33}}
              
              \put(252,85){$\epsilon(t)$}
              \put(250,95){\vector(0,-1){20}}
              \put(251,60){\circle{24}}
              \put(247,58){$\Large+$}
              
              \put(265,66){$g(t)$}
              \put(265,60){\vector(1,0){20}}
              \put(0,20){\siga}
\end{picture}
}

\def\siga{\setlength{\unitlength}{0.46pt}
\begin{picture}(290,108)
\thinlines    \put(10,26){\vector(1,0){260}}
              \put(16,12){\vector(0,1){88}}
              \put(48,27){\vector(0,1){34}}
              \put(79,27){\vector(0,1){54}}
              \put(130,27){\vector(0,1){46}}
              \put(188,26){\vector(0,1){40}}
              \put(262,14){$t$}
              \put(182,16){$\red{t_K}$}
              \put(195,63){$\red{a_K}$}
              \put(125,15){$\red{t_k}$}
              \put(137,68){$\red{a_k}$}
              \put(75,14){$\red{t_2}$}
              \put(84,73){$\red{a_2}$}
              \put(43,15){$\red{t_1}$}
              \put(54,56){$\red{a_1}$}
              \multiput(150,46)(8,0){3}{\circle*{4}}
\end{picture}
}

\def\modele{\tt\setlength{\unitlength}{0.92pt}
\begin{picture}(250,106)
\thinlines    \put(20,70){$s(t)=\sum_n \rfn \delta(t-n\Delta t)$}
              \put(105,60){\vector(1,0){50}}
              \put(154,52){\framebox(50,16){$\Nc(t|0,v)$}}
              
              \put(205,75){$g_{\rfb}(t)$}
              \put(204,60){\vector(1,0){33}}
              
              \put(252,85){$\epsilon(t)$}
              \put(250,95){\vector(0,-1){20}}
              \put(251,60){\circle{24}}
              \put(247,58){$\Large+$}
              
              \put(265,66){$g(t)$}
              \put(265,60){\vector(1,0){20}}
              \put(0,20){\sigb}
\end{picture}
}

\def\sigb{\setlength{\unitlength}{0.46pt}
\begin{picture}(290,108)
\thinlines    \put(10,26){\vector(1,0){260}}
              \put(10,12){\vector(0,1){88}}
              
              \put(20,27){\vector(0,1){34}}
              \put(40,27){\vector(0,1){14}}
              \put(60,27){\vector(0,1){54}}
              \put(80,27){\vector(0,1){31}}
              \put(100,27){\vector(0,1){46}}
              \put(120,27){\vector(0,1){40}}
              \put(140,27){\vector(0,1){48}}
              \put(160,27){\vector(0,1){34}}
              \put(180,27){\vector(0,1){14}}
              \put(200,27){\vector(0,1){31}}
              \put(220,27){\vector(0,1){46}}
              \put(240,27){\vector(0,1){40}}
              
              \put(270,14){$t$}
              \put(15,10){$1$}
              \put(15,63){$\red{f_1}$}
              \put(135,10){$n$}
              \put(135,68){$\red{f_n}$}
              \put(235,10){$N$}
              \put(235,68){$\red{f_N}$}

\end{picture}
}

\def\sigbr{\setlength{\unitlength}{0.46pt}
\begin{picture}(290,108)
\thinlines    \put(10,26){\vector(1,0){260}}
              \put(10,12){\vector(0,1){88}}
              
              \put(20,27){\vector(0,1){34}}
              \put(40,27){\vector(0,1){14}}
              \put(60,27){\vector(0,1){54}}
              \put(80,27){\vector(0,1){31}}
              \put(100,27){\vector(0,1){46}}
              \put(120,27){\vector(0,1){40}}
              \put(140,27){\vector(0,1){48}}
              \put(160,27){\vector(0,1){34}}
              \put(180,27){\vector(0,1){14}}
              \put(200,27){\vector(0,1){31}}
              \put(220,27){\vector(0,1){46}}
              \put(240,27){\vector(0,1){40}}
              \put(270,14){$t$}
              \put(15,10){$1$}
              \put(15,63){$\red{f_1}$}
              \put(135,10){$n$}
              \put(135,68){$\red{f_n}$}
              \put(235,10){$N$}
              \put(235,68){$\red{f_N}$}

              \put(60,10){$\red{\Delta t}$}
\end{picture}
}

\def\modelf{\tt\setlength{\unitlength}{0.92pt}
\begin{picture}(250,106)
\thinlines    \put(20,70){$s(t)=\sum_n \rfn \delta(t-n\red{\Delta t})$}
              \put(105,60){\vector(1,0){50}}
              \put(154,52){\framebox(50,16){$\Nc(t|0,\red{v})$}}
              
              \put(205,75){$g_{\rfb,\rthetab}(t)$}
              \put(204,60){\vector(1,0){33}}
              
              \put(252,85){$\epsilon(t)$}
              \put(250,95){\vector(0,-1){20}}
              \put(251,60){\circle{24}}
              \put(247,58){$\Large+$}
              
              \put(265,66){$g(t)$}
              \put(265,60){\vector(1,0){20}}
              \put(0,20){\sigbr}
\end{picture}
}

\def\sig{\tt\setlength{\unitlength}{1.38pt}
\begin{picture}(290,108)
\thinlines    
              \put(10,26){\vector(1,0){260}}
              \put(48,27){\vector(0,1){34}}
              \put(16,12){\vector(0,1){88}}
              \put(79,27){\vector(0,1){54}}
              \put(102,27){\vector(0,1){31}}
              \put(130,27){\vector(0,1){46}}
              \put(188,26){\vector(0,1){40}}
              \put(262,14){$t$}
              \put(22,90){$g(t)$}
              \put(182,16){$t_K$}
              \put(195,63){$a_K$}
              \put(125,15){$t_k$}
              \put(137,68){$a_k$}
              \put(75,14){$t_2$}
              \put(84,73){$a_2$}
              \put(43,15){$t_1$}
              \put(54,56){$a_1$}
              \multiput(150,46)(8,0){3}{\circle*{4}}
\end{picture}
}

\def\decgb{\tt\setlength{\unitlength}{0.92pt}
\begin{picture}(363,122)
\thinlines    \put(248,66){\vector(1,0){14}}
              \put(215,57){\framebox(33,18){$h(t)$}}
              \put(288,66){\vector(1,0){22}}
              \put(270,64){$\large\Sigma$}
              \put(275,67){\circle{26}}
              \put(275,102){\vector(0,-1){20}}
              \put(280,94){$b(t)$}
              \put(296,75){$y(t)$}
              \multiput(100,35)(0,8){3}{\circle*{4}}
              \put(194,72){$x(t)$}
              \put(193,66){\vector(1,0){23}}
              \put(176,61){$\large\Sigma$}
              \put(181,65){\circle{26}}
              \put(80,92){\framebox(43,18){$a_1$}}
              \put(80,58){\framebox(43,18){$a_2$}}
              \put(79,10){\framebox(43,18){$a_K$}}
              \put(123,66){\vector(1,0){45}}
              \put(123,101){\vector(4,-3){44}}
              \put(122,18){\vector(1,1){44}}
              \put(55,101){\vector(1,0){23}}
              \put(55,66){\vector(1,0){23}}
              \put(55,19){\vector(1,0){23}}
              \put(10,104){$g_1(t-t_1)$}
              \put(10,69){$g_2(t-t_2)$}
              \put(10,22){$g_K(t-t_k)$}
\end{picture}
}

\def\decgc{\tt\setlength{\unitlength}{0.92pt}
\begin{picture}(287,68)
\thinlines    \put(36,20){\vector(1,0){23}}
              \put(101,20){\vector(1,0){23}}
              \put(167,20){\vector(1,0){16}}
              \put(124,12){\framebox(43,18){$h(t)$}}
              \put(210,20){\vector(1,0){23}}
              \put(192,20){$\large\Sigma$}
              \put(196,23){\circle{26}}
              \put(196,57){\vector(0,-1){20}}
              \put(200,49){$b(t)$}
              \put(218,30){$y(t)$}
              \put(15,20){$s(t)$}
              \put(101,32){$x(t)$}
              \put(59,11){\framebox(43,18){$g(t)$}}
\end{picture}
}

\def\sigd{
\begin{picture}(290,108)
\thinlines    \put(10,26){\vector(1,0){260}}
              \put(48,27){\vector(0,1){34}}
              \put(16,12){\vector(0,1){88}}
              \put(79,27){\vector(0,1){54}}
              \put(102,27){\vector(0,1){31}}
              \put(130,27){\vector(0,1){46}}
              \put(188,26){\vector(0,1){40}}
              \put(262,14){$t$}
              \put(22,90){$g(t)$}
              \put(182,16){$t_K$}
              \put(195,63){$a_K$}
              \put(125,15){$t_k$}
              \put(137,68){$a_k$}
              \put(75,14){$t_2$}
              \put(84,73){$a_2$}
              \put(43,15){$t_1$}
              \put(54,56){$a_1$}
              \multiput(150,46)(8,0){3}{\circle*{4}}
\end{picture}
}

\def\wt#1{\widetilde{#1}}
\def\Thetab{\bm{\Theta}}
\def\Thetabh{\widehat{\Thetab}}
\def\abh{\widehat{\ab}}
\def\kbt{\tilde{\kb}}
\def\kt{\tilde{k}}
\def\gammat{\tilde{\gamma}}
\def\Sigmat{\tilde{\Sigma}}
\def\etat{\tilde{\eta}}

\def\rHb{\red{\Hb}}
\def\rFb{\red{\Fb}}

\def\wt#1{\widetilde{#1}}
\def\Thetab{\bm{\Theta}}
\def\Thetabh{\widehat{\Thetab}}
\def\abh{\widehat{\ab}}
\def\kbt{\tilde{\kb}}
\def\kt{\tilde{k}}
\def\gammat{\tilde{\gamma}}
\def\Sigmat{\tilde{\Sigma}}
\def\etat{\tilde{\eta}}
\def\zetat{\tilde{\zeta}}
\def\Mt{\tilde{M}}
\def\Mh{\hat{M}}
\def\Nt{\tilde{N}}
\def\atk{{\tilde{a}_k}}
\def\ztk{{\tilde{z}_k}}

\def\rh{\red{h}}
\def\rvbe{\rvb_{\epsilon}}
\def\rvei{\rv_{\epsilon_i}}
\def\rVbe{\rVb_{\epsilon}}
\def\rVbeh{\red{\Vbh}_{\epsilon}}
\def\vbh{\hat{\vb}}
\def\rVbfh{\red{\Vbh}_f}
\def\rvbeh{\rvbh_{\epsilon}}
\def\rvbet{\rvbt_{\epsilon}}
\def\rvbfh{\rvbh_f}

\def\rvbf{\rvb_f}
\def\rvfj{\rv_{f_j}}
\def\rVbf{\red{\Vb}_f}

\def\alphaez{\alpha_{\epsilon_0}}
\def\betaez{\beta_{\epsilon_0}}
\def\alphafz{\alpha_{f_0}}
\def\betafz{\beta_{f_0}}

\def\wh#1{\widehat{#1}}
\def\disp#1{{\displaystyle #1}}

\def\ER{\mbox{I\kern-.25em R}}
\def\EC{\mbox{C\kern-.8em C}}
\def\EZ{\mbox{Z\kern-.55em Z}}
\def\EN{\mbox{N\kern-.8em N}}

\def\bm#1{\mbox{\boldmath $#1$}}
\def\sb{{\bm s}}
\def\ub{{\bm u}}
\def\xb{{\bm x}}
\def\yb{{\bm y}}
\def\db{{\bm{d}}}
\def\Ab{{\bm A}}
\def\Bb{{\bm B}}
\def\Fb{{\bm F}}
\def\Rb{{\bm R}}
\def\Sb{{\bm S}}
\def\Ub{{\bm U}}

\def\thetab{\bm{\theta}}
\def\phib{\bm{\phi}}
\def\lambdab{\bm{\lambda}}
\def\Lambdab{\bm{\Lambda}}
\def\Sigmab{\bm{\Sigma}}
\def\lra{\longrightarrow}
\def\intg{\int\kern-1.1em\int}
\def\intd{\int\kern-.8em\int}

\def\apriori{{\em a priori} }
\def\aposteriori{{\em a posteriori} }

\def\d#1{\,\mbox{d}#1}
\def\esp#1{\mbox{E}\left\{ #1 \right\}}
\def\espx#1#2{\mbox{E}_{#1}\left\{ #2 \right\}}
\def\expf#1{\exp\left[ {#1} \right]}
\def\dpdx#1#2{\frac{\partial #1}{\partial #2}}
\def\dpdxd#1#2{\frac{\partial^2 #1}{\partial #2^2}}

\def\ent#1{\hbox{H}\left[#1\right]}
\def\dent#1{\delta\hbox{H}\left[#1\right]}
\def\rent#1{\hbox{D}\left[#1\right]}
\def\kullback#1{\hbox{K}\left[#1\right]}
\def\pent#1{\hbox{PE}\left[#1\right]}
\def\mdiv#1{\hbox{J}\left[#1\right]}
\def\infmut#1{\hbox{I}\left[#1\right]}
\def\idisc#1{\hbox{ID}\left[#1\right]}
\def\j{\mbox{j}}

\def\pxtheta{p(x | \thetab)}
\def\pxthetas{p(x | \thetab^*)}
\def\pxthetasd{p(x | \thetab^*+\Delta\thetab)}

\def\tr#1{\hbox{tr}\left(#1\right)}
\def\det#1{\hbox{det}\left(#1\right)}

\def\ieeeSSC{IEEE Transactions on Systems Science and Cybernetics}
\def\ieeeSP{IEEE Transactions on Signal Processing}			
\def\ieeeP{Proceedings of the IEEE}					
\def\ieeeIT{IEEE Transactions on Information Theory}			
\def\TS{Traitement du Signal}						
\def\siamCO{SIAM Journal of Control}					
\def\AISM{Annals of Institute of Statistical Mathematics}		
\def\ieeeASSP{IEEE Transactions on Acoustics Speech and Signal Processing}
\def\ieeeAC{IEEE Transactions on Automatic and Control}			

\def\jan{janvier\xspace}
\def\feb{f\'evrier\xspace}
\def\mar{mars\xspace}
\def\apr{avril\xspace}
\def\may{mai\xspace}
\def\jun{juin\xspace}
\def\jul{juillet\xspace}
\def\aug{ao\^ut\xspace}
\def\sep{septembre\xspace}
\def\oct{octobre\xspace}
\def\nov{novembre\xspace}
\def\dec{d\'ecembre\xspace}
\def\Jan{January\xspace}	
\def\Feb{February\xspace}
\def\Mar{March\xspace}
\def\Apr{April\xspace}
\def\May{May\xspace}
\def\Jun{June\xspace}
\def\Jul{July\xspace}
\def\Aug{August\xspace}
\def\Sep{September\xspace}
\def\Oct{October\xspace}
\def\Nov{November\xspace}
\def\Dec{December\xspace}

\def\espq#1{\left<#1\right>_q}

\def\rZbe{\rZb_{\epsilon}}
\def\nubt{\widetilde{\nub}}

\def\rVbf{\rVb_{f}}
\def\rvbf{\rvb_{f}}
\def\rvfj{\red{v}_{f_j}}
\def\rvbhe{\rvbh_{\epsilon}}
\def\rvbhf{\rvbh_{f}}
\def\bsplit{\begin{split}}
\def\esplit{\end{split}}
\def\alphaei{\alpha_{\epsilon_i}}
\def\alphafj{\alpha_{f_j}}
\def\betaei{\beta_{\epsilon_i}}
\def\betafj{\beta_{f_j}}

\def\bgbh{\widehat{\bgb}}
\def\rfh{\widehat{\rf}}
\def\rFh{\widehat{\rF}}
\def\rc{\red{r}}
\def\rch{\widehat{\rc}}
\def\partialx{\frac{\partial }{\partial x}}
\def\partialy{\frac{\partial }{\partial y}}
\def\partialr{\frac{\partial }{\partial r}}
\def\partialxd{\frac{\partial^2 }{\partial x^2}}
\def\partialyd{\frac{\partial^2 }{\partial y^2}}
\def\partialrd{\frac{\partial^2 }{\partial r^2}}
\def\cosphi{\cos(\phi)}
\def\sinphi{\sin(\phi)}
\def\sumk{\sum_{k=1}^K}
\def\sumj{\sum_{j=1}^N}
\def\sumi{\sum_{i=1}^M}
\def\summ{\sum_{m=1}^M}
\def\sumn{\sum_{n=1}^N}
\def\summn{\sum_{m=1}^M\sum_{n=1}^N}

\def\df{\dot{f}}
\def\ddf{\ddot{f}}

\def\rf{\red{f}}
\def\rdf{\red{\df}}
\def\rddf{\red{\ddf}}

\def\rfb{\red{\fb}}
\def\rdfb{\red{\dot{\fb}}}
\def\rddfb{\red{\ddot{\fb}}}

\def\bdg{\blue{\dot{g}}}
\def\bddg{\blue{\ddot{g}}}

\def\bgb{\blue{\gb}}
\def\bdg{\blue{\dot{g}}}

\def\bdgb{\blue{\dot{\gb}}}
\def\bddgb{\blue{\ddot{\gb}}}

\def\epsilonbd{\dot{\epsilonb}}
\def\epsilonbdd{\ddot{\epsilonb}}

\def\rdfbh{\widehat{\rdfb}}
\def\rddfbh{\widehat{\rddfb}}

\def\rqh{\red{\widehat{q}}}
\def\rdfh{\red{\widehat{\dot{f}}}}
\def\rddfh{\red{\widehat{\ddot{f}}}}

\def\vek{v_{\epsilon_k}}
\def\rveih{\widehat{\rve}_i}
\def\rvfjh{\widehat{\rvf}_j}

\def\alphaxiz{{\alpha_{\xi_0}}}
\def\betaxiz{{\beta_{\xi_0}}}
\def\alphazz{{\alpha_{z_0}}}

\def\betazz{{\beta_{z_0}}}
\def\rVbz{{\rVb_z}}
\def\rvbz{{\rvb_z}}
\def\rvxi{{\rv_\xi}}
\def\rvzj{{\rv_{z_j}}}
\def\rvxih{{\widehat{\rv}_\xi}}
\def\rvbzh{{\widehat{\rvb}_z}}
\def\rvzjh{{\widehat{\rv}_{z_j}}}

\def\alphae{{\alpha_\epsilon}}

\def\alphaet{{\widetilde{\alpha}_\epsilon}}
\def\alphaft{{\widetilde{\alpha}_f}}
\def\betae{{\beta_\epsilon}}
\def\betaet{{\widetilde{\beta}_\epsilon}}
\def\betaft{{\widetilde{\beta}_f}}
\def\rfih{\hat{\rf}_i}
\def\rfbk{\rfb^{(k)}}
\def\rfbkp{\rfb^{(k+1)}}

\def\rmh{\hat{\red{m}}}

\def\rmk{{\red{m}_k}}
\def\rmkh{{\hat{\red{m}}_k}}
\def\rvk{{\red{v}_k}}
\def\rvkh{{\hat{\red{v}}_k}}
\def\rak{{\red{a}_k}}
\def\rakh{{\hat{\red{a}}_k}}

\def\rvh{\hat{\red{v}}}
\def\rveh{\hat{\rve}}

\def\rmt{\tilde{\red{m}}}
\def\rvt{\tilde{\red{v}}}
\def\rvet{\tilde{\rve}}
\def\rveih{\red{\hat{v}}_{\epsilon_i}}

\def\rveb{\red{\vb_{\epsilon}}}
\def\rvebh{\red{\vbh_{\epsilon}}}

\def\blocgi{\begin{picture}(80,50)
  \put(60,40){\makebox(0,0){$\red{m}$}}
  \put(60,40){\circle{20}}
  \put(60,30){\vector(0,-1){20}}
  \put(60,0){\circle{20}}
  \put(60,0){\makebox(0,0){$\blue{g}_i$}}
  \put(20,0){\circle{20}}
  \put(20,0){\makebox(0,0){$\epsilon_i$}}
  \put(30,0){\vector(1,0){20}}
\end{picture}
}
\def\blocgfi{\begin{picture}(80,50)
  \put(60,40){\makebox(0,0){$\rf_i$}}
  \put(60,40){\circle{20}}
  \put(60,30){\vector(0,-1){20}}
  \put(60,0){\circle{20}}
  \put(60,0){\makebox(0,0){$\blue{g}_i$}}
  \put(20,0){\circle{20}}
  \put(20,0){\makebox(0,0){$\epsilon_i$}}
  \put(30,0){\vector(1,0){20}}
\end{picture}
}
\def\blocghfi{\begin{picture}(80,50)
  \put(60,40){\makebox(0,0){$\rf_i$}}
  \put(60,40){\circle{20}}
  \put(60,30){\vector(0,-1){20}}
  \put(65,20){\makebox(0,0){$h$}}
  \put(60,0){\circle{20}}
  \put(60,0){\makebox(0,0){$\blue{g}_i$}}
  \put(20,0){\circle{20}}
  \put(20,0){\makebox(0,0){$\epsilon_i$}}
  \put(30,0){\vector(1,0){20}}
\end{picture}
}
\def\blocgie{\begin{picture}(80,50)
  \put(60,40){\makebox(0,0){$\red{m}$}}
  \put(60,40){\circle{20}}
  \put(60,30){\vector(0,-1){20}}
  \put(60,0){\circle{20}}
  \put(60,0){\makebox(0,0){$\blue{g}_i$}}
  \put(30,0){\circle{20}}
  \put(30,0){\makebox(0,0){$\epsilon_i$}}
  \put(40,0){\vector(1,0){10}}
  \put(0,0){\circle{20}}
  \put(0,0){\makebox(0,0){$\rve$}}
  \put(10,0){\vector(1,0){10}}
\end{picture}
}
\def\blocgievei{\begin{picture}(80,50)
  \put(60,40){\makebox(0,0){$\red{m}$}}
  \put(60,40){\circle{20}}
  \put(60,30){\vector(0,-1){20}}
  \put(60,0){\circle{20}}
  \put(60,0){\makebox(0,0){$\blue{g}_i$}}
  \put(30,0){\circle{20}}
  \put(30,0){\makebox(0,0){$\epsilon_i$}}
  \put(40,0){\vector(1,0){10}}
  \put(0,0){\circle{20}}
  \put(0,0){\makebox(0,0){$\rvei$}}
  \put(10,0){\vector(1,0){10}}
\end{picture}
}
\def\blocfgie{\begin{picture}(80,100)
  \put(60,40){\makebox(0,0){$\rf_i$}}
  \put(60,40){\circle{20}}
  \put(60,30){\vector(0,-1){20}}
  \put(65,20){\makebox(0,0){$\blue{h}$}}
  
  \put(60,0){\circle{20}}
  \put(60,0){\makebox(0,0){$\blue{g}_i$}}
  \put(30,0){\circle{20}}
  \put(30,0){\makebox(0,0){$\epsilon_i$}}
  \put(40,0){\vector(1,0){10}}
  \put(0,0){\circle{20}}
  \put(0,0){\makebox(0,0){$\rve$}}
  \put(10,0){\vector(1,0){10}}
\end{picture}
}
\def\blocvfgie{\begin{picture}(80,100)
  \put(60,80){\makebox(0,0){$\rv_i$}}
  \put(60,80){\circle{20}}
  \put(60,70){\vector(0,-1){20}}
  
  \put(60,40){\makebox(0,0){$\rf_i$}}
  \put(60,40){\circle{20}}
  \put(60,30){\vector(0,-1){20}}
  \put(65,20){\makebox(0,0){$\blue{h}$}}
  
  \put(60,0){\circle{20}}
  \put(60,0){\makebox(0,0){$\blue{g}_i$}}
  \put(30,0){\circle{20}}
  \put(30,0){\makebox(0,0){$\epsilon_i$}}
  \put(40,0){\vector(1,0){10}}
  \put(0,0){\circle{20}}
  \put(0,0){\makebox(0,0){$\rve$}}
  \put(10,0){\vector(1,0){10}}
\end{picture}
}
\def\blocvfhgie{\begin{picture}(80,100)
  \put(60,80){\makebox(0,0){$\rv_i$}}
  \put(60,80){\circle{20}}
  \put(60,70){\vector(0,-1){20}}
  
  \put(60,40){\makebox(0,0){$\rf_i$}}
  \put(60,40){\circle{20}}
  \put(60,30){\vector(0,-1){20}}
  
  \put(30,30){\circle{20}}
  \put(30,30){\makebox(0,0){$\red{h}$}}
  \put(38,22){\vector(1,-1){15}}
  
  \put(60,0){\circle{20}}
  \put(60,0){\makebox(0,0){$\blue{g}_i$}}
  \put(30,0){\circle{20}}
  \put(30,0){\makebox(0,0){$\epsilon_i$}}
  \put(40,0){\vector(1,0){10}}
  \put(0,0){\circle{20}}
  \put(0,0){\makebox(0,0){$\rve$}}
  \put(10,0){\vector(1,0){10}}
\end{picture}
}
\def\blocvfhgiei{\begin{picture}(80,100)
  \put(60,80){\makebox(0,0){$\rv_i$}}
  \put(60,80){\circle{20}}
  \put(60,70){\vector(0,-1){20}}
  
  \put(60,40){\makebox(0,0){$\rf_i$}}
  \put(60,40){\circle{20}}
  \put(60,30){\vector(0,-1){20}}
  
  \put(30,30){\circle{20}}
  \put(30,30){\makebox(0,0){$\red{h}$}}
  \put(38,22){\vector(1,-1){15}}
  
  \put(60,0){\circle{20}}
  \put(60,0){\makebox(0,0){$\blue{g}_i$}}
  \put(30,0){\circle{20}}
  \put(30,0){\makebox(0,0){$\epsilon_i$}}
  \put(40,0){\vector(1,0){10}}
  \put(0,0){\circle{20}}
  \put(0,0){\makebox(0,0){$\rvei$}}
  \put(10,0){\vector(1,0){10}}
\end{picture}
}
\def\blocfgHe{\begin{picture}(80,100)
  \put(60,40){\makebox(0,0){$\rfb$}}
  \put(60,40){\circle{20}}
  \put(60,30){\vector(0,-1){20}}
  \put(65,20){\makebox(0,0){$\blue{\Hb}$}}
  
  \put(60,0){\circle{20}}
  \put(60,0){\makebox(0,0){$\bgb$}}
  \put(30,0){\circle{20}}
  \put(30,0){\makebox(0,0){$\epsilonb$}}
  \put(40,0){\vector(1,0){10}}
  \put(0,0){\circle{20}}
  \put(0,0){\makebox(0,0){$\rve$}}
  \put(10,0){\vector(1,0){10}}
\end{picture}
}
\def\blocvfgHe{\begin{picture}(80,100)
  \put(60,80){\makebox(0,0){$\rvb$}}
  \put(60,80){\circle{20}}
  \put(60,70){\vector(0,-1){20}}
  
  \put(60,40){\makebox(0,0){$\rfb$}}
  \put(60,40){\circle{20}}
  \put(60,30){\vector(0,-1){20}}
  \put(65,20){\makebox(0,0){$\bHb$}}
  
  \put(60,0){\circle{20}}
  \put(60,0){\makebox(0,0){$\bgb$}}
  \put(30,0){\circle{20}}
  \put(30,0){\makebox(0,0){$\epsilonb$}}
  \put(40,0){\vector(1,0){10}}
  \put(0,0){\circle{20}}
  \put(0,0){\makebox(0,0){$\rve$}}
  \put(10,0){\vector(1,0){10}}
\end{picture}
}
\def\blocvfgHei{\begin{picture}(80,100)
  \put(60,80){\makebox(0,0){$\rvb$}}
  \put(60,80){\circle{20}}
  \put(60,70){\vector(0,-1){20}}
  
  \put(60,40){\makebox(0,0){$\rfb$}}
  \put(60,40){\circle{20}}
  \put(60,30){\vector(0,-1){20}}
  \put(65,20){\makebox(0,0){$\bHb$}}
  
  \put(60,0){\circle{20}}
  \put(60,0){\makebox(0,0){$\bgb$}}
  \put(30,0){\circle{20}}
  \put(30,0){\makebox(0,0){$\epsilonb$}}
  \put(40,0){\vector(1,0){10}}
  \put(0,0){\circle{20}}
  \put(0,0){\makebox(0,0){$\rvei$}}
  \put(10,0){\vector(1,0){10}}
\end{picture}
}
\def\blocvfHge{\begin{picture}(80,100)
  \put(60,80){\makebox(0,0){$\rvb$}}
  \put(60,80){\circle{20}}
  \put(60,70){\vector(0,-1){20}}
  
  \put(60,40){\makebox(0,0){$\rfb$}}
  \put(60,40){\circle{20}}
  \put(60,30){\vector(0,-1){20}}
  
  \put(30,30){\circle{20}}
  \put(30,30){\makebox(0,0){$\rHb$}}
  \put(38,22){\vector(1,-1){15}}
  
  \put(60,0){\circle{20}}
  \put(60,0){\makebox(0,0){$\bgb$}}
  \put(30,0){\circle{20}}
  \put(30,0){\makebox(0,0){$\epsilonb$}}
  \put(40,0){\vector(1,0){10}}
  \put(0,0){\circle{20}}
  \put(0,0){\makebox(0,0){$\rve$}}
  \put(10,0){\vector(1,0){10}}
\end{picture}
}
\def\blocvfHgei{\begin{picture}(80,100)
  \put(60,80){\makebox(0,0){$\rvb$}}
  \put(60,80){\circle{20}}
  \put(60,70){\vector(0,-1){20}}
  
  \put(60,40){\makebox(0,0){$\rfb$}}
  \put(60,40){\circle{20}}
  \put(60,30){\vector(0,-1){20}}
  
  \put(30,30){\circle{20}}
  \put(30,30){\makebox(0,0){$\rHb$}}
  \put(38,22){\vector(1,-1){15}}
  
  \put(60,0){\circle{20}}
  \put(60,0){\makebox(0,0){$\bgb$}}
  \put(30,0){\circle{20}}
  \put(30,0){\makebox(0,0){$\epsilonb$}}
  \put(40,0){\vector(1,0){10}}
  \put(0,0){\circle{20}}
  \put(0,0){\makebox(0,0){$\rvei$}}
  \put(10,0){\vector(1,0){10}}
\end{picture}
}
\def\blocvhFge{\begin{picture}(80,100)
  \put(60,80){\makebox(0,0){$\rvb$}}
  \put(60,80){\circle{20}}
  \put(60,70){\vector(0,-1){20}}
  
  \put(60,40){\makebox(0,0){$\rhb$}}
  \put(60,40){\circle{20}}
  \put(60,30){\vector(0,-1){20}}
  
  \put(30,30){\circle{20}}
  \put(30,30){\makebox(0,0){$\bFb$}}
  \put(38,22){\vector(1,-1){15}}

  \put(60,0){\circle{20}}
  \put(60,0){\makebox(0,0){$\bgb$}}
  \put(30,0){\circle{20}}
  \put(30,0){\makebox(0,0){$\epsilonb$}}
  \put(40,0){\vector(1,0){10}}
  \put(0,0){\circle{20}}
  \put(0,0){\makebox(0,0){$\rve$}}
  \put(10,0){\vector(1,0){10}}
\end{picture}
}
\def\blocvhFgei{\begin{picture}(80,100)
  \put(60,80){\makebox(0,0){$\rvb$}}
  \put(60,80){\circle{20}}
  \put(60,70){\vector(0,-1){20}}
  
  \put(60,40){\makebox(0,0){$\rhb$}}
  \put(60,40){\circle{20}}
  \put(60,30){\vector(0,-1){20}}
  
  \put(30,30){\circle{20}}
  \put(30,30){\makebox(0,0){$\bFb$}}
  \put(38,22){\vector(1,-1){15}}

  \put(60,0){\circle{20}}
  \put(60,0){\makebox(0,0){$\bgb$}}
  \put(30,0){\circle{20}}
  \put(30,0){\makebox(0,0){$\epsilonb$}}
  \put(40,0){\vector(1,0){10}}
  \put(0,0){\circle{20}}
  \put(0,0){\makebox(0,0){$\rvei$}}
  \put(10,0){\vector(1,0){10}}
\end{picture}
}
\def\blochFge{\begin{picture}(80,40)
  \put(60,40){\makebox(0,0){$\rhb$}}
  \put(60,40){\circle{20}}
  \put(60,30){\vector(0,-1){20}}
  
  \put(30,30){\circle{20}}
  \put(30,30){\makebox(0,0){$\bFb$}}
  \put(38,22){\vector(1,-1){15}}

  \put(60,0){\circle{20}}
  \put(60,0){\makebox(0,0){$\bgb$}}
  \put(30,0){\circle{20}}
  \put(30,0){\makebox(0,0){$\epsilonb$}}
  \put(40,0){\vector(1,0){10}}
  \put(0,0){\circle{20}}
  \put(0,0){\makebox(0,0){$\rve$}}
  \put(10,0){\vector(1,0){10}}
\end{picture}
}
\def\blochFgei{\begin{picture}(80,40)
  \put(60,40){\makebox(0,0){$\rhb$}}
  \put(60,40){\circle{20}}
  \put(60,30){\vector(0,-1){20}}
  
  \put(30,30){\circle{20}}
  \put(30,30){\makebox(0,0){$\bFb$}}
  \put(38,22){\vector(1,-1){15}}

  \put(60,0){\circle{20}}
  \put(60,0){\makebox(0,0){$\bgb$}}
  \put(30,0){\circle{20}}
  \put(30,0){\makebox(0,0){$\epsilonb$}}
  \put(40,0){\vector(1,0){10}}
  \put(0,0){\circle{20}}
  \put(0,0){\makebox(0,0){$\rvei$}}
  \put(10,0){\vector(1,0){10}}
\end{picture}
}
\def\blocfHge{\begin{picture}(80,40)  
  \put(60,40){\makebox(0,0){$\bfb$}}
  \put(60,40){\circle{20}}
  \put(60,30){\vector(0,-1){20}}
  
  \put(30,30){\circle{20}}
  \put(30,30){\makebox(0,0){$\rHb$}}
  \put(38,22){\vector(1,-1){15}}
  
  \put(60,0){\circle{20}}
  \put(60,0){\makebox(0,0){$\bgb$}}
  \put(30,0){\circle{20}}
  \put(30,0){\makebox(0,0){$\epsilonb$}}
  \put(40,0){\vector(1,0){10}}
  \put(0,0){\circle{20}}
  \put(0,0){\makebox(0,0){$\rve$}}
  \put(10,0){\vector(1,0){10}}
\end{picture}
}
\def\blocfHgei{\begin{picture}(80,40)  
  \put(60,40){\makebox(0,0){$\bfb$}}
  \put(60,40){\circle{20}}
  \put(60,30){\vector(0,-1){20}}
  
  \put(30,30){\circle{20}}
  \put(30,30){\makebox(0,0){$\rHb$}}
  \put(38,22){\vector(1,-1){15}}
  
  \put(60,0){\circle{20}}
  \put(60,0){\makebox(0,0){$\bgb$}}
  \put(30,0){\circle{20}}
  \put(30,0){\makebox(0,0){$\epsilonb$}}
  \put(40,0){\vector(1,0){10}}
  \put(0,0){\circle{20}}
  \put(0,0){\makebox(0,0){$\rvei$}}
  \put(10,0){\vector(1,0){10}}
\end{picture}
}
\def\blocFgHe{\begin{picture}(80,40)
  \put(60,40){\makebox(0,0){$\bfb$}}
  \put(60,40){\circle{20}}
  \put(60,30){\vector(0,-1){20}}
  \put(65,20){\makebox(0,0){$\blue{\rHb}$}}
  
  \put(60,0){\circle{20}}
  \put(60,0){\makebox(0,0){$\bgb$}}
  \put(30,0){\circle{20}}
  \put(30,0){\makebox(0,0){$\epsilonb$}}
  \put(40,0){\vector(1,0){10}}
  \put(0,0){\circle{20}}
  \put(0,0){\makebox(0,0){$\rve$}}
  \put(10,0){\vector(1,0){10}}
\end{picture}
}

\def\ftrfi{\bg(r,\phi)}
\def\wth{\widehat{\tilde{h}}}
\def\varx#1#2{\mbox{Var}_{#1}\left\{#2\right\}}
\def\rvbxi{\rvb_{\xi}}
\def\rVbxi{\rVb_{\xi}}
\def\rvbxih{\rvbh_{\xi}}

\def\repsilon{\red{\epsilon}}

\def\vzj{v_{z_j}}
\def\vxij{v_{\xi_j}}
\def\vei{v_{\epsilon_i}}

\def\vzjh{\widehat{v}_{z_j}}
\def\vxijh{\widehat{v}_{\xi_j}}
\def\veih{\widehat{v}_{\epsilon_i}}

\def\vbeh{\widehat{\vb}_{\epsilon}}
\def\vbxih{\widehat{\vb}_{\xi}}
\def\vbzh{\widehat{\vb}_{z}}

\def\Vbeh{\widehat{\Vb}_{\epsilon}}
\def\Vbxih{\widehat{\Vb}_{\xi}}
\def\Vbzh{\widehat{\Vb}_{z}}

\def\Ybeh{\widehat{\Yb}_{\epsilon}}
\def\Ybxih{\widehat{\Yb}_{\xi}}
\def\Ybzh{\widehat{\Yb}_{z}}

\def\vf{v_f}
\def\vbe{\vb_{\epsilon}}
\def\vbz{\vb_z}
\def\vxi{v_{\xi}}
\def\vbxi{\vb_{\xi}}
\def\Vbe{\Vb_{\epsilon}}
\def\Vbxi{\Vb_{\xi}}
\def\Vbz{\Vb_z}

\def\half{\frac{1}{2}}

\def\dpdx#1#2{\frac{\partial {#1}}{\partial {#2}}}
\def\dpdxd#1#2{\frac{\partial^2 {#1}}{\partial {#2}^2}}
\def\dpdxdy#1#2#3{{{\partial ^2 {#1}}\over{\partial {#2} \partial {#3}}}}

\def\KL#1#2{\mbox{KL}\left[#1 : #2\right]}
\def\Covx#1#2{\mbox{Cov}_{#1}(#2)}
\def\braket#1#2{<{#1}(#2)>}

\def\rlambdab{\red{\lambdab}}
\def\bve{\blue{\ve}}
\def\ralphafjt{\red{\tilde{\alpha}_{f_j}}}
\def\rbetafjt{\red{\tilde{\beta}_{f_j}}}
\def\rSigmabt{\red{\tilde{\Sigmab}}}
\def\rfjt{\red{\tilde{f_j}}}
\def\zetab{\bm{\zeta}}
\def\bvxi{\blue{v_\xi}}
\def\bgamma{\blue{\gamma}}
\def\rgamma{\red{\gamma}}
\def\alphagamz{\alpha_{\gamma_0}}
\def\betagamz{\beta_{\gamma_0}}
\def\alphazetaz{\alpha_{\zeta_0}}
\def\betazetaz{\beta_{\zeta_0}}

\def\bvf{\blue{v_f}}
\def\bvzeta{\blue{v_\zeta}}
\def\rgb{\red{\gb}}
\def\rvxii{\red{v_{\xi_i}}}
\def\rvzeta{\red{v_{\zeta}}}

\usepackage{tikz,mathpazo}
\usetikzlibrary{shapes.geometric, arrows}
\usetikzlibrary{positioning,shapes,shadows,arrows}
\usetikzlibrary{mindmap,backgrounds}
\tikzstyle{startstop} = [rectangle, rounded corners, minimum width=1.5cm, minimum height=0.5cm,text centered, draw=black]
\tikzstyle{io} = [trapezium, trapezium left angle=70, trapezium right angle=110, minimum width=1.5cm, minimum height=0.6cm, text centered, draw=black]
\tikzstyle{process} = [rectangle, minimum width=1cm, minimum height=0.5cm, text centered, draw=black]
\tikzstyle{note} = [text centered]
\tikzstyle{decision} = [diamond, minimum width=1.5cm, minimum height=0.3cm, text centered, draw=black]
\tikzstyle{arrow} = [thick,->,>=stealth]

\def\blambda{\blue{\lambda}}
\def\blambdab{\blue{\lambdab}}

\def\fxy{f(x,y)}
\def\afxy{a\left(f(x,y)\right)}

\def\hxy{h(x,y)}
\def\gxy{g(x,y)}

\def\gt{\widetilde{g}}
\def\ft{\widetilde{f}}
\def\htxy{\widetilde{h}(x,y)}
\def\htzero{\widetilde{h}_0}
\def\htone{\widetilde{h}_1}
\def\httwo{\widetilde{h}_2}

\def\hdxy{h_d(x,y)}
\def\gdxy{g_d(x,y)}

\def\exy{\epsilon(x,y)}
\def\nxy{n(x,y)}
\def\rxy{r(x,y)}

\def\dxy{d(x,y)}
\def\dhxy{d_h(x,y)}
\def\dfxy{d_f(x,y)}

\def\fhxy{\hat{f}(x,y)}
\def\ghxy{\hat{f}(x,y)}
\def\ehxy{\hat{\epsilon}(x,y)}
\def\nhxy{\hat{n}(x,y)}

\def\ftuv{\widetilde{f}(u,v)}
\def\htuv{\widetilde{h}(u,v)}
\def\gtuv{\widetilde{g}(u,v)}
\def\rtuv{\widetilde{r}(u,v)}
\def\ntuv{\widetilde{n}(u,v)}
\def\dtuv{\widetilde{d}(u,v)}
\def\dhtuv{\widetilde{d}_h(u,v)}
\def\dftuv{\widetilde{d}_f(u,v)}

\def\htcuv{\widetilde{h}^*(u,v)}
\def\ftcuv{\widetilde{f}^*(u,v)}

\def\ftuv{\widetilde{f}(u,v)}
\def\htuv{\widetilde{h}(u,v)}
\def\etuv{\widetilde{\epsilon}(u,v)}

\def\ftcuv{\widetilde{f}^*(u,v)}
\def\htcuv{\widetilde{h}^*(u,v)}
\def\etcuv{\widetilde{\epsilon}^*(u,v)}

\def\fhtuv{\widehat{\widetilde{f}}(u,v)}
\def\hhtuv{\widehat{\widetilde{h}}(u,v)}

\def\fhtxy{\widehat{\widetilde{f}}(x,y)}
\def\hhtxy{\widehat{\widetilde{h}}(x,y)}

\def\fhtcuv{\widehat{\widetilde{f}}^*(u,v)}
\def\hhtcuv{\widehat{\widetilde{h}}^*(u,v)}
\def\ghtcuv{\widehat{\widetilde{g}}^*(u,v)}

\def\htzerouv{\widetilde{h}_0(u,v)}
\def\htoneuv{\widetilde{h}_1(u,v)}
\def\httwouv{\widetilde{h}_2(u,v)}
\def\htzerocuv{\widetilde{h}_0^*(u,v)}
\def\htonecuv{\widetilde{h}_1^*(u,v)}
\def\httwocuv{\widetilde{h}_2^*(u,v)}
\def\gtoneuv{\widetilde{g}_1(u,v)}
\def\gttwouv{\widetilde{g}_2(u,v)}

\def\ftuvk{\widetilde{f}^{(k)}(u,v)}
\def\htuvk{\widetilde{h}^{(k)}(u,v)}

\def\fhxyk{\widehat{f}^{(k)}(x,y)}
\def\hhxyk{\widehat{h}^{(k)}(x,y)}
\def\fhxykp{\widehat{f}^{(k+1)}(x,y)}
\def\hhxykp{\widehat{h}^{(k)}(x,y)}

\def\fhk{\widehat{f}^{(k)}}
\def\hhk{\widehat{h}^{(k)}}
\def\fhkp{\widehat{f}^{(k+1)}}
\def\hhkp{\widehat{h}^{(k)}}

\def\fxyk{f^{(k)}(x,y)}
\def\fxykp{f^{(k+1)}(x,y)}

\def\fkxy{f^{(k)}(x,y)}
\def\fkpxy{f^{(k+1)}(x,y)}

\def\ftuvkp{\widetilde{f}^{(k+1)}(u,v)}
\def\htuvkp{\widetilde{h}^{(k+1)}(u,v)}

\def\hktuv{\widetilde{h}_k(u,v)}
\def\gktuv{\widetilde{g}_k(u,v)}

\def\bfxy{\blue{f}(x,y)}
\def\bhxy{\blue{h}(x,y)}
\def\bgxy{\blue{g}(x,y)}

\def\rfxy{\red{f}(x,y)}
\def\rhxy{\red{h}(x,y)}

\def\rftuv{\widetilde{\red{f}}(u,v)}
\def\rhtuv{\widetilde{\red{h}}(u,v)}

\def\bgtuv{\widetilde{\blue{g}}(u,v)}

\def\alphabh{\widehat{\alphab}}

\def\dfdx#1#2{\frac{\partial #1}{\partial #2}}

\def\zxy{z(x,y)}
\def\Txy{T(x,y)}
\def\fxy{f(x,y)}
\def\gzxy{g_0(x,y)}
\def\gxy{g(x,y)}
\def\phixy{\phi(x,y)}
\def\exy{e(x,y)}
\def\epsxy{\epsilon(x,y)}

\def\zhxy{\hat{z}(x,y)}
\def\Thxy{\hat{T}(x,y)}
\def\gzhxy{\hat{g}_0(x,y)}
\def\fhxy{\hat{f}(x,y)}
\def\phihxy{\hat{\phi}(x,y)}
\def\epshxy{\hat{\epsilon}(x,y)}